\documentclass{m2an}
\usepackage{enumerate}

\usepackage{amssymb}
\usepackage{braket, amsmath, bm, amsthm, url}
\usepackage[linesnumbered,ruled,vlined]{algorithm2e}
\usepackage{graphicx}

\newtheorem{corollary}{Corollary}[section]
\newtheorem{definition}{Definition}[section]
\newtheorem{remark}{Remark}[section]

\begin{document}
\newpage
\setcounter{page}{1}
\title{Sensitivity of quantum PageRank}
\author{Hirotada Honda}\address{
Faculty of Information Networking for Innovation and Design, Toyo University\\
1-7-11 Akabanedai 115-0053, Japan\\
E-mail: honda012@toyo.jp
Phone:+81-3-5924-2658}
%
\makeatletter
\@addtoreset{equation}{section}
\def\theequation{\thesection.\arabic{equation}}
\date{...}
\begin{abstract} 
In this paper, we discuss the sensitivity of quantum PageRank.
By using the finite dimensional perturbation theory, we estimate the change of the quantum PageRank under a small analytical perturbation on the Google matrix. In addition, we will show the way to estimate the lower bound of the convergence radius as well as the error bound of the finite sum in the expansion of the perturbed PageRank.
 \end{abstract}
%
%
\subjclass{Key words:Quantum walk, PageRank, perturbation theory.}
%
%
\maketitle
%
\section{Introduction}
Quantum information has recently attracted attention.
Especially, quantum walks have been actively discussed from both the theoretical and practical viewpoints~\cite{Nielsen_2011}. 

A major breakthrough was the quadratic speedup in the search algorithms of quantum walks~\cite{Grover_1996}. The paper \cite{Szegedy_2004} extended the random walk concept on a graph to quantum walks on a directed graph.

Based on the ideas presented in~\cite{Ambainis_2004} and \cite{Watrous_2001}, he introduced an unitary operator that works on a Hilbert space whose elements are made of 
pairs of edges and the components of the state transition matrix. Recently, studies on the mixing time of the quantum walk on the graph has also been accelerated~\cite{Aharonov_2001}. 

In parallel with these researches, the ranking process has been actively discussed in classic network studies. A prominent ranking process is Google's PageRank algorithm, which appropriately sorts web pages in order of their importance and impact.  

Quantum walk research on PageRank was pioneered in~\cite{Paparo_2012}, who proposed a quantum version of the PageRank algorithm. Based on Szegedy's ideas~\cite{Szegedy_2004}, they quantized the behavior of a random walker modeled in the Google matrix and defined the unitary operator. They derived some interesting relations between the quantum and original versions of PageRank. For instance, the eigenvalues and eigenvectors of the unitary operator in the quantum PageRank can be represented by those of a matrix composed of the Google-matrix elements. They also numerically computed some characteristics of quantum PageRank and revealed its similarities to and differences from the classical PageRank.

Previously, we studied the behavior of the classical PageRank subjected to small perturbations.
Because the Perron--Frobenius theorem ensures good characteristics of the principal eigenvalue of the Google matrix, the PageRank could be analytically perturbed by analytically perturbing the Google matrix.
However, the perturbations of the quantum PageRank are less tractable because all the eigenvalues of the unitary operator have unity magnitude.
Therefore, we should consider the behavior of all eigenvalues and their corresponding eigenvectors under a perturbation.
To overcome these difficulties, we will exploit the characteristics of normal operators, the reduction process, and the transformation functions in analytical perturbation theory. 

The remainder of this study is organized as follows. The next section introduces the terms and notations used throughout this study. Section 3 reviews the previous arguments concerning classical and quantum PageRanks. Our main results are presented in Section 4 and are proven in Section 5. The conclusions are presented in Section 6. 

\section{Terms and notations}
 In this study, $N$ denotes an arbitrary natural number, and the vectors are labeled as $\ket{\psi}$.
The tensor product of two spaces ${\bf C}^N$ is denoted as ${\bf C}^N \otimes {\bf C}^N$, and its elemts are
$\ket{\phi} \otimes \ket{\psi}$ or simply $\ket{\phi , \psi}$. 
The transposed row vector of $\ket{\psi}$ is $\bra{\psi}$, and the inner product of two column vectors
$\ket{\phi}$ and $\ket{\psi}$ is denoted as $\braket{\phi | \psi}$.

A general complex number and its conjugate are denoted as $z$ and $z^*$, respectively ($z, z^* \in {\bm C}$ ).
${\rm Re}(z)$ and ${\rm Im}(z)$ are the real and imaginary parts of $z \in {\bf C}$, respectively.
The imaginary unit is ${\rm i}=\sqrt{-1}$.

The $p$-norm of a general vector   ${\bm u} = (u_1,u_2, \ldots , u_N)^{\rm T}$ is defined as 
\begin{align*}
& \| {\bm u} \|_p \equiv 
  \Bigl( 
    \sum_{j=1}^N |x_j|^p
  \Bigr)^\frac{1}{p} \quad (p \in [1,\infty)),
\end{align*}
and 
$\|{\bm u}\|_\infty \equiv \max_{1 \leq j \leq N} |u_j|.$
For a squared matrix ${\bm M} = [m_{ij}] \in {\bf M}_{N \times N}$
(where ${\bf M}_{N \times N}$ is a set of $N$ times $N$ matrices), we define 
\begin{align*}
&\bigl\| {\bm M} \bigr\|_\infty = \max_{1 \leq i \leq N} \sum_{j=1}^N |m_{ij}|, 
  \quad
  \bigl\| {\bm M} \bigr\|_1 = \max_{1 \leq j \leq N} \sum_{i=1}^N |m_{ij}|.
\end{align*}
We also define the {\it operator norm}
$
\|{\bm M}\| \equiv \sup_{ x \ne 0} \frac{\|{\bm M} x\|}{\|x\|},
$

where the norm on the right-hand side is the usual norms of vector spaces: 
$\|x\|_p$. Additionally, we simply denote the vector norm $\|\cdot \|_2$ by $\| \cdot \|$ and apply the corresponding operator norms. 
For a matrix ${\bm M}$, we define the set of eigenvalues, resolvent set, trace, and spectral radius as 
$\sigma({\bm M})$, $P(T)$,
${\rm tr}({\bm M})$,
 and ${\rm spr}( {\bm M})$, respectively. That is,
$$
{\rm spr}( {\bm M}) = \lim_{n \rightarrow \infty} \| {\bm M}^n\|^\frac{1}{n}.
$$
(See for instance \cite{Kato_1980}).  The multiplicity of $\lambda \in \sigma({\bm M})$ is denoted as $m(\lambda)$.
When $m(\lambda)=1$, $\lambda$ is simple.
We also define $R({\bm M})$ as the range of the maps represented by ${\bm M}$.

For a point $x \in {\bf R}^N$ and a general set $S \subset {\bf R}^N$, we have  
${\rm dist}(x,S) \equiv \min_{y \in S} |x-y|$, where $|x-y|$ is the usual Euclidean distance 
between two points $x$ and $y$ in ${\bm R}^N$. 
In an unitary space $H$ with its inner product $(\cdot, \cdot),$ if an operator ${\bm M}^*$ satisfies
$\bigl( {\bm M}^*x,y \bigr)=\bigl( x, {\bm M}y \bigr) \quad \forall x, \; y \in H,$ then
${\bm M}^*$ is called as the {\it adjoint operator} of ${\bm M}$.
An operator ${\bm M}$ is called {\it normal} if it commutes with its conjugate operator:
${\bm M}{\bm M}^*={\bm M}^*{\bm M}$.
The term {\it isolation distance} of an eigenvalue $\lambda_h$ 
of a matrix ${\bm M}$ means the distance between $\lambda_h$
 and the other eigenvalues of ${\bm M}$, that is, ${\rm dist}(\lambda_h,\sigma({\bm M}))$.
It is often convenient to set $\|{\bm \pi}\|_1 = 1$ for a PageRank vector ${\bm \pi},$
but Tikhonov's theorem~\cite{Batkai_2017} states that norms in a finite-dimensional vector space are equivalent. 
The $r$-th derivative of a general function $f$ with a single argument is sometimes denoted as $D^r f \; (r \in {\bf N})$.
\section{Background}
\subsection{Google matrix and PageRank}
The PageRank algorithm was originally proposed in \cite{Brin_and_Page_1998}. 
Many researchers have subsequently made computational and theoretical contributions to this algorithm. Although thorough sensitivity analyses of the PageRank vector exist in the literature~\cite{Langville_2006}, we consider that more general situations will appear in practical situations, which warrant further discussion.
The author's previous paper \cite{Honda_2018} presented a sensitivity analysis of PageRank under small analytic perturbations.
We now briefly define a Google matrix and overview its characteristics.
The Internet is modeled as a directed graph $G(V,E)$, where $V$ and $E$ are the sets of nodes and edges, respectively. 
The notation $N=\# V$ describes the number of nodes in this graph. For later use, the elements of $V$ are labeled as $1,2,\ldots,N$.

The row-stochastic matrix ${\bm W}$ is defined by ${\bm W} = {\bm H} + \frac{\bm a}{N}{\bm e}^{\rm T}$, where ${\bm a}=(a_j)$.
We set $a_j=1$ if web page $j$ is a {\it dangling node} (a node without any outgoing links) and 
$a_j=0$ otherwise~\cite{Langville_2006}. 
%
 The {\it hyperlink matrix} ${\bm H} =[h_{ij}]$ is a weighted adjacency matrix defined by
$h_{ij} =1/D_i$ if there is a link from web page  $i$ to web page  $j$, and $h_{ij}=0$ otherwise, where $D_i$ is the number of
outgoing links from node $i$.
The Google matrix ${\bm G}$ is defined by
$
{\bm G} = \alpha {\bm W} + \frac{(1-\alpha)}{N} {\bm e}{\bm v}^{\rm T},
$
where ${\bm e}$ is a column vector whose elements are all $1$, and ${\bm v}$ is a 
non-negative vector satisfying $\|{\bm v}\|_1=1$ (called the {\it personalization vector}). The constant $\alpha \in (0,1)$, is called the {\it damping factor}. The quantity
$(1-\alpha)$ represents the 
probability that an Internet surfer navigates from one web page $(i)$ to another $(j)$.
By definition, ${\bm G}$ is completely dense, primitive, and irriducible~\cite{Langville_2006}.

The following lemma is the foundation of Google matrix theory. 
\begin{lmm}
The Google matrix ${\bm G}$ has a simple principal eigenvalue $\eta_1=1$. All eigenvalues $\{\eta_n\}_n$ of ${\bm G}$ satisfy
$
\eta_1 = 1 > | \eta_2 |  \geq |\eta_3| \geq \ldots.
$
Additionally, there exists a non-negative left-eigenvector ${\bm \pi}$ corresponding to the principal eigenvalue $\eta_1=1$,
which is known as the PageRank vector.
\end{lmm}

\subsection{Quantum PageRank}
The PageRank vector denotes the probability that a random walker on the Internet stays on each node during the stationary state. Recently, this concept has been replaced by the quantum walk. To our knowledge, quantum walk on a directed graph was introduced in \cite{Szegedy_2004}.
Based on this idea, the authors of \cite{Paparo_2012} studied the quantum PageRank. They started their arguments by introducing vectors of the form: 
\begin{align*}
&\ket{\psi_j} \equiv \ket{j}_1 \otimes \sum_{k=1}^N \sqrt{g_{jk}} \ket{k}_2
 \quad (j=1,2,\ldots, N), 
\end{align*}
where $g_{jk}$ is the $(j,k)$ element of the Google matrix ${\bm G}$.
Here, the indices $\ket{ \cdot }_l \; (l=1,2)$ show the start and end nodes of the edges, respectively. It is easily seen that
\begin{align*}
&\braket{\psi_j | \psi_k} = \delta_{jk} \quad \forall j,k,
\end{align*}
where $\delta_{jk}$ is the Kronecker's delta.
Therefore, $\{ \ket{\psi_j} \}_j$ forms an orthonormal basis in a subspace of ${\mathcal H} \equiv {\bf C}^N \otimes {\bf C}^N$.
The authors of \cite{Paparo_2012} also defined the operators
\begin{align*}
&{\bm B} \equiv \sum_{j=1}^N \ket{\psi_j}\bra{\psi_j}, \quad
{\bm U} \equiv {\bm S}_w \bigl( 2 {\bm B} - {\bm I}\bigr),
\end{align*}
where ${\bm I}$ is the $N$-dimensional squared unit matrix, and ${\bm S}_w$ is the swap operator:
\begin{align*}
&{\bm S}_w \equiv \sum_{j,k=1}^N \ket{k,j} \bra{j,k}.
\end{align*}
Clearly, ${\bm U}$ is an unitary operator on ${\mathcal H}$~\cite{Paparo_2012}. 
Now, let us denote its eigenvalues and corresponding normalized eigenvectors as
$\{ \mu \}$ and $\{ \ket{ \mu} \}$, respectively. Note that because ${\bm U}$ is 
unitary, $\{ \ket{\mu} \}$ forms an orthonormal basis in ${\mathcal H}$.

Now, let us think of the subspace $H_d$ spanned by $\{ \ket{\psi_j } \}_j$ and 
$\{ {\bm S}_w \ket{\psi_j} \}_j$~\cite{Paparo_2012}. We also consider its orthogonal subspace $H_d^{\perp}$, on which ${\bm U}$ acts as $-{\bm S}_w$ whose eigenvalues 
are just $\pm1$.

The eigenvalues of ${\bm U}$ (other than those of ${\bm S}_w$) are then expressed in the form
$
\mu = \lambda \pm {\rm i} \sqrt{1-\lambda^2},
$
where the eigenvalues $\lambda$ of the positive symmetric matrix ${\bm T} = [ t_{ij}]$ lie in $[0,1]$~\cite{Szegedy_2004}, and
$t_{ij} = \sqrt{g_{ij}g_{ji}} $
(see also \cite{Szegedy_2004}).  Hereafter, the number of eigenvalues of ${\bm T}$ (excluding multiple values) is denoted as $s$.
That is, $\sigma({\bm T}) = \{\lambda_h\}_{h=1}^s.$
Now, the subspace of ${\mathcal H}$ which is spanned by the eigenvectors of ${\bm U}$
corresponding to $\mu = \lambda \pm {\rm i} \sqrt{1-\lambda^2}$ is denoted as
$H_e$ hereafter. 

The quantum PageRank for node $i \in V$ at time $m \in {\bf N}$ is then defined as
\begin{align}
&I_q(i,m | \psi(0) ) = 
 \Bigl|
  \sum_{\mu \in H_e, \atop j \in V} \mu^{2m} \braket{j,i | \mu} \braket{\mu | \psi(0)}
 \Bigr|^2 \notag \\
&\hspace{19mm}
 \equiv 
 \bigl| N_q( i,m |\psi(0)) \bigr|^2,
\end{align}
where $\psi(0)$ is the initial state of the walk, which lies in the subspace 
$H_d$ spanned by $\{ \ket{\psi_j } \}_j$ and  $\{ {\bm S}_w \ket{\psi_j} \}_j$~\cite{Paparo_2012}. In addition, the notation $\mu \in H_e$ under the summation symbol
implies that we take the sum over eigenvalues whose corresponding eigenvectors
belong to $H_e$. The definition (3.1) implies that we consider the spectrum decomposition 
of ${\bm U}$ on the space $H_e$, that is, we project the space $H_d$ onto $H_e$.
\begin{remark}
The authors of {\rm \cite{Paparo_2012}} used the notation 
\begin{align*}
&I_q(P_i,m) = \left\|
 \sum_{\mu} \mu^{2m} \;_2 \braket{i | \mu} \braket{\mu | \psi(0)}
\right\|^2,
\end{align*}
which does not clarify the meaning of the projection. In addition, as the contents are scalar values, we adopt the absolute-value notation here. 
\end{remark}
\subsection{Formulation and settings}
In this paper, we impose a small perturbation on the Google matrix ${\bm G} \in {\bf M}_{N \times N}$.
The perturbed Google matrix is denoted as ${\bm G}(\chi) = [g_{ij}(\chi)]$, where $\chi \in {\bf C}$.
Likewise, the perturbed ${\bm W}$ is denoted as ${\bm W}(\chi)$. The perturbation is presumed not to remove any nodes or edges, and to preserve the characteristics of
 ${\bm G}$ and ${\bm W}$, both of which are row-stochastic, irreducible, and primitive. 
In this article, we impose an analytic perturbation on the Google matrix, as discussed in~\cite{Honda_2018}:
\begin{align}
&{\bm G}(\chi) = {\bm G} + \chi {\bm G}^{(1)} + \chi^2  {\bm G}^{(2)} + \ldots.
\end{align}
We assume that 
\begin{align}
&\|{\bm G}^{(l)} \| \leq A_0B_0^{l-1} \quad l=1,2,\ldots.
\end{align}
where $A_0$ and $B_0$ are positive constants independent of $l$. Assuming that (3.2) converges, such constants certainly exist. 
We also define the matrix ${\bm T}(\chi) = [t_{ij}(\chi)]$ with 
$t_{ij}(\chi) = \sqrt{ g_{ij}(\chi) g_{ji}(\chi)}$, and
$\ket{ \psi_j(\chi) } = \ket{j}_1 \otimes \sum_{k=1}^N \sqrt{g_{jk}(\chi)}\ket{k}_2$.

Then, ${\bm T}(\chi)$ is holomorphic with respect to $\chi$ when $|\chi|$ is sufficiently small, and can be expanded as an infinite sum:
\begin{align*}
&{\bm T}(\chi) = {\bm T} + \sum_{l=1}^\infty \chi^l {\bm T}^{(l)}.
\end{align*}
The corresponding eigenvalues of the perturbed ${\bm T}(\chi)$ are denoted as $\{ \lambda(\chi)\}$:
obviously, $\lambda(0)=\lambda$ for each eigenvalue. To introduce the quantum walk, we further define 
$$
{\bm B}(\chi) = \sum_{j=1}^N \ket{\psi_j (\chi)} \bra{ \psi_j (\chi)}, \quad
{\bm U}(\chi) = {\bm S}_w \Bigl( 2{\bm B}(\chi) - {\bm I} \Bigr).
$$
The corresponding eigenvalues and normalized eigenvectors of the perturbed ${\bm U}(\chi)$ are denoted as $\{ \mu(\chi)\}$ and $\bigl\{ \ket{\mu(\chi)} \bigr\}$, respectively. 
As the above-defined ${\bm U}$ is an unitary operator on ${\mathcal H}$, all of its eigenvalues have unity magnitude.
The quantum PageRank at time $m$ under perturbation is then found by
\begin{align*}
&I_q^{(\chi)}(i,m | \psi(0) ) = \left|
 \sum_{\mu \in H_e, \atop j \in V} (\mu(\chi))^{2m}
 \braket{j,i | \mu (\chi)} \braket{\mu(\chi) | \psi(0)}
\right|^2,
\end{align*}
where the summation is taken over the set of corresponding unperturbed $\mu$'s
 in $H_e$. The perturbation is assumed to be very small, such that
$$
|\mu(\chi)| < 1 \quad \forall \mu=\mu(0) \in H_e. 
$$
%
\section{Main results}
We now give the main statement in this article.
\begin{thrm}
Let $m \in {\bf N}$ be arbitrary, and assume that there exists at least one dangling node in the network. Then, when $\chi \in {\bf C}$ is sufficiently small, 
the temporal quantum PageRank is holomorphic with respect to $\chi$ on the real axis. That is, within a neighborhood of the real axis, it can be represented in the form:
\begin{align}
&I_q^{(\chi)}(i,m |\psi(0)) = I_q(i,m|\psi(0))
  + \sum_{n=1}^\infty \chi^l I_q^{(n)} (i,m |\psi(0)) \notag \\
&\hspace{40mm}
 \forall i=1,\ldots,N, \; 
  \forall \psi(0) \in H_d.
\end{align}
Under Assumption {\rm (3.3)}, the convergence radius of this expansion is determined by $N$, $m$, $A_0$, $B_0$, the elements of ${\bm G}$, and the isolation distance of ${\bm T}$.
In addition, each term in the right-hand side of {\rm (4.1)} is estimated from above in the region stated as above: 
\begin{align*}
&|I_q^{(n)} (i,m |\psi(0)) | \leq A_1B_1^{n-1} \; (n=1,2,\ldots),
\end{align*}
where the constants $A_1, \; B_1>0$ depend on the same quantities as the convergence radius.
\end{thrm}
\begin{remark}
The term `holomorphic on the real axis' means that the quantum PageRank is holomorphic in a neighborhood of the real axis.
\end{remark}
$I_q(i,m|\psi(0))$ is oscillatory, and and is not temporally convergent in general. Instead, the following quantity is convergent: 

\begin{definition}
The average temporal quantum PageRank is defined as:
$$
\bar{I}_q(t,i|\psi(0)) \equiv \frac{1}{t} \sum_{m=0}^{t-1} I_q(i,m | \psi(0) )
 \quad \forall t \in {\bf N}.
$$
\end{definition}

In this definition, we can state the following facts. 
\begin{lmm}
The average of the temporal quantum PageRank converges as time tends to infinity. The limit is given by
\begin{align*}
&I_\infty(i | \psi(0) ) \equiv \lim_{t \rightarrow \infty}\bar{I}_q(t,i | \psi(0)) \\
&\hspace{17mm}
 = \sum_{p,q} \sum_{j \in V}
     \braket{\mu_p | \psi(0)} \braket{ \mu_q | \psi(0)}^*
      \braket{j,i | \mu_p }  \braket{ \mu_q | j,i},
\end{align*}
where the first summation is taken over the pairs of eigenvalues $(\mu_p,\mu_q)$ satisfying $\mu_p=\mu_q$.
\end{lmm}

\begin{corollary}
If all eigenvalues of ${\bm T}$ are simple, then 
\begin{align*}
&I_\infty(i | \psi(0) )
  = \sum_{h=1}^N\sum_{j \in V}  | \braket{ \mu_h | \psi(0)} |^2 | \braket{ j,i | \mu_h} |^2
\end{align*}
\end{corollary}
This statement directly derives from the results of \cite{Aharonov_2001}.
They also discussed the upper bound of the mixing time:

\begin{lmm}
The following inequality holds: 
\begin{align}
&|\bar{I}_q(t,i| \psi(0)) - I_\infty (i| \psi(0))|
  \leq \sum_{\mu_p, \mu_q \in H_e \atop \mu_p \ne \mu_q}
  \frac{2 | \braket{ \mu_p | \psi(0)}|^2 }{ t |\mu_p - \mu_q|}.
\end{align}
\end{lmm}

It is seen that 
 $\bar{I}_q(t,i | \psi(0)) $ and $I_\infty(i | \psi(0) )$ are not usually analytic under an analytic perturbation on ${\bm G}$.
Actually (as seen later), some eigenvalues of ${\bm U}(\chi)$ may split into several eigenvalues as $\chi$ grows.

 In that case, 
the terms in the summation on the right-hand side of (4.2) will increase, meaning that they may change drastically under even a small perturbation.
\section{Proof of Theorem 4.1}
This section will prove Theorem 4.1. The proof is divided into several steps. 

\subsection{Perturbation on ${\bm G}$}
Assuming an analytic perturbation on ${\bm G}$, we define
$$
g_{ij}(\chi) = g_{ij} + \chi g_{ij}^{(1)} + \chi^2 g_{ij}^{(2)} + \ldots \equiv g_{ij} + \tilde{g}_{ij}(\chi).
$$

The elements of ${\bm T}(\chi) = [t_{ij}(\chi)]$ are then given as
\begin{align*}
&t_{ij}(\chi) 
 = \sqrt{ \bigl( g_{ij} + \tilde{g}_{ij}(\chi) \bigr)  \bigl(g_{ji} + \tilde{g}_{ij}(\chi) \bigr)} 
%
 = t_{ij} + \sum_{n=1}^\infty \chi^n t_{ij}^{(n)},
\end{align*}
for small $\chi $ that satisfies $|\tilde{g}_{ij}(\chi)| < 1$.  Note that each element of the Google matrix differs from zero~\cite{Langville_2006}, and  $t_{ij}^{(n)}$ is defined as 
$$
t_{ij}^{(n)} = \frac{1}{n!} \Bigl( \frac{{\rm d}}{{\rm d} \chi} \Bigr)^n t_{ij}(\chi) \Bigr|_{\chi=0},
$$
(see~\cite{Knopp_1945} for instance). By Leibnitz's rule, we have 
\begin{align*}
&
\Bigl( \frac{{\rm d}}{{\rm d} \chi} \Bigr)^n t_{ij}(\chi)  = 
 t_{ij} \sum_{k=0}^n \;_nC_k
  \Bigl( \frac{{\rm d}}{{\rm d} \chi} \Bigr)^k
   \Bigl[
    \sqrt{1+\breve{g}_{ij}(\chi) }
     \Bigr] 
%
  \Bigl( \frac{{\rm d}}{{\rm d} \chi} \Bigr)^{n-k}
   \Bigl[
    \sqrt{1+\breve{g}_{ji}(\chi) }
     \Bigr],
\end{align*}
where 
$$
\breve{g}_{ij}(\chi)
  \equiv \frac{\tilde{g}_{ij}(\chi)}{g_{ij}} \equiv \frac{ g_{ij}(\chi) - g_{ij}}{g_{ij}} .
$$
(Note that $\breve{g}_{ij}(0) = \tilde{g}_{ij}(0) = 0$). 
By the derivative formula of composite functions~\cite{McKiernan_1956}, and setting $f(z)=\sqrt{1+z}$, we have
\begin{align*}
&\Bigl( \frac{{\rm d}}{{\rm d} \chi} \Bigr)^n
    \Bigl[
     \sqrt{1+\breve{g}_{ij}(\chi) }
      \Bigr] 
%
 = \Bigl( \frac{{\rm d}}{{\rm d} \chi} \Bigr)^n
    f(\breve{g}_{ij}(\chi)) \\
&\hspace{34mm} 
= \sum_{r=1}^n \frac{D^r f(\breve{g}_{ij} (\chi))}{r!} 
%
 \sum_{p_1+\ldots + p_r=n \atop p_i \geq 1}
   \frac{n!}{p_1! p_2! \ldots p_r!} 
    \bigl( D^{p_1} \breve{g}_{ij}\bigr)
     \bigl( D^{p_2} \breve{g}_{ij}\bigr) \ldots 
      \bigl( D^{p_r} \breve{g}_{ij} \bigr).
\end{align*}
Noting that
\begin{align*}
&\Bigl( \frac{{\rm d}}{{\rm d} \chi} \Bigr)^r   \Bigl[ \sqrt{1+z} \Bigr]
 = \;_\frac12 C_r\bigl( 1+z \bigr)^{-\frac{(2r-1)}{2}},
\end{align*}
and
 $D^{p_i} \breve{g}_{ij}(\chi)\Bigr|_{\chi=0} = \frac{ p_i! g_{ij}^{(p_i)}}{ g_{ij}} \; ( i=1,2,\ldots,r)$, we have
\begin{align*}
&\Bigl( \frac{{\rm d}}{{\rm d} \chi} \Bigr)^n   \Bigl[
    \sqrt{1+\breve{g}_{ij}(\chi) }
 \Bigr] \Bigr|_{ \chi=0}
=
 n! \sum_{r=1}^n \frac{\;_\frac12 C_r}{ g_{ij}^r}
      \sum_{p_1+\ldots + p_r=n \atop p_i\geq 1} \prod_{\tau=1}^r g_{ij}^{(p_\tau)}.
\end{align*}
In terms of this expression, $t_{ij}^{(n)}$ becomes
\begin{align}
&t_{ij}^{(n)} = 
 t_{ij} \Biggl[
  \sum_{r^\prime=1}^{n} \frac{\;_\frac12 C_{r^\prime}}{g_{ji}^{r^\prime}} 
   \sum_{p_1+\ldots + p_{r^\prime}=n \atop p_i \geq 1}  \prod_{\tau^\prime=1}^{r^\prime}
     g_{ji}^{(p_{\tau^\prime})}
    \Biggr] \notag \\
&\hspace{10mm}
 +t_{ij} \sum_{k=1}^{n-1} 
 \Biggl[
  \sum_{r=1}^k \frac{\;_\frac12 C_r}{g_{ij}^r} 
   \sum_{p_1+\ldots + p_r=k \atop p_i \geq 1}  \prod_{\tau=1}^r g_{ij}^{(p_\tau)}
    \Biggr] 
%
\Biggl[
  \sum_{r^\prime=1}^{n-k} \frac{\;_\frac12 C_{r^\prime}}{g_{ji}^{r^\prime}} 
   \sum_{p_1+\ldots + p_{r^\prime}=n-k \atop p_i \geq 1}  \prod_{\tau^\prime=1}^{r^\prime}
     g_{ji}^{(p_{\tau^\prime})}
    \Biggr] \notag \\
&\hspace{10mm}
+ t_{ij} \Biggl[
  \sum_{r=1}^{n} \frac{\;_\frac12 C_r}{g_{ij}^r} 
   \sum_{p_1+\ldots + p_r=n \atop p_i \geq 1}  \prod_{\tau=1}^r
     g_{ij}^{(p_{\tau})}
    \Biggr] \notag \\
&\hspace{5mm}
 \equiv I_{1(n)}^{(i,j)}+I_{2(n)}^{(i,j)}+I_{3(n)}^{(i,j)}.
\end{align}
For instance, we have 
\begin{align*}
&t_{ij}^{(1)} 
 = \frac{1}{2t_{ij}} \Bigl( g_{ij}^{(1)} g_{ji} + g_{ji}^{(1)} g_{ij}  \Bigr).
\end{align*}

\subsection{Perturbation on ${\bm T}$}
Based on the discussion in the previous subsection, we represent
\begin{align*}
&{\bm T}(\chi) = {\bm T} + \sum_{n=1}^\infty \chi^n {\bm T}^{(n)} \quad
 (|\tilde{g}_{ij} (\chi)| < 1 \; \forall i,j \in V).
\end{align*}

Note that ${\bm T}$ and ${\bm T}(\chi)$ for $\chi \in {\bf R}$ are symmetric, and consequently semisimple.
The resolvent of ${\bm T}$ is denoted as 
$$
{\bm R}(\zeta) \equiv ({\bm T} -\zeta {\bm I} )^{-1},
$$
and the eigenprojection corresponding to eigenvalue $\lambda_h$ is 
$$
{\bm P}_h \equiv -\frac{1}{2\pi {\rm i}} 
 \int_{\Gamma_h} {\bm R}(\zeta) \; {\rm d}\zeta.
$$
Here, $\Gamma_h$ is an arbitrary convex loop enclosing $\lambda_h$ in the complex plain, excluding all other eigenvalues of ${\bm T}$.
We now represent the resolvent and eigenprojection under a small perturbation. We denote the resolvent of ${\bm T}(\chi)$ by
${\bm R}(\zeta,\chi) \equiv \bigl( {\bm T}(\chi) - \zeta {\bm I}\bigr)^{-1}$,
where $\zeta \in {\bf C} \bigcap \sigma({\bm T}(\chi))^c$. 
${\bm R}(\zeta,\chi)$ is defined for all $\zeta \notin \sigma \bigl({\bm T}(\chi)\bigr)$. We know that  $R(\zeta,\chi)$ is holomorphic with 
respect to $\zeta$ and $\chi$ for such a $\zeta$ (see Theorem II.1.5 in~\cite{Kato_1980}),
and is expanded as follows: 
\begin{align}
&{\bm R}(\zeta,\chi) 
 = {\bm R}(\zeta) \bigl[ {\bm I} + {\bm A}(\chi) {\bm R}(\zeta) \bigr]^{-1} 
%
 = {\bm R}(\zeta) + \sum_{l=1}^\infty \chi^l {\bm R}^{(l)}(\zeta),
\end{align}
where ${\bm A}(\chi) = {\bm T}(\chi) - {\bm T}$, and 
\begin{align*}
&{\bm R}^{(l)}(\zeta) \equiv
  \sum_{\nu_1+\ldots + \nu_p=l \atop \nu_j \geq 1}
 (-1)^p {\bm R}(\zeta) {\bm T}^{(\nu_1)}
            {\bm R}(\zeta) {\bm T}^{(\nu_2)} \ldots
%
    {\bm R}(\zeta) {\bm T}^{(\nu_p)}  {\bm R}(\zeta). 
\end{align*}
Here,  the summation on the right-hand side is taken over all possible values of $p \in {\bm N}$ and $(\nu_1, \ldots, \nu_p)$ meeting the condition below the summation symbol.
Integrating the infinite series (5.2) over $\Gamma_h$, the perturbation of the projection operator ${\bm P}_h(\chi)$ is rendered as
\begin{align}
&{\bm P}_h(\chi) = -\frac{1}{2\pi {\rm i}} \int_{\Gamma_h}
   {\bm R}(\zeta,\chi) \; {\rm d}\zeta
    = {\bm P}_h + \sum_{l=1}^\infty \chi^l {\bm P}_h^{(l)}.
\end{align}
${\bm P}_h(0) ={\bm P}_h 
 = -\frac{1}{2\pi {\rm i}} \int_{\Gamma_h} {\bm R}(\zeta) \; {\rm d}\zeta.
$
In (5.3), ${\bm P}_h^{(l)}$ is given by
\begin{align}
&{\bm P}_h^{(l)}  
 = - \frac{1}{2\pi {\rm i}}  \sum_{\nu_1+\ldots + \nu_p=l, \atop \nu_k \geq 1}
%
   (-1)^p \int_{\rm \Gamma_h} 
    {\bm R}(\zeta)  {\bm T}^{(\nu_1)}
     {\bm R}(\zeta)  {\bm T}^{(\nu_2)} \ldots
      {\bm R}(\zeta)  {\bm T}^{(\nu_p)}
       {\bm R}(\zeta)  \; {\rm d}\zeta.
\end{align}
We also define
$$
{\bm S}_h =  -\frac{1}{2\pi {\rm i}}
 \int_{\Gamma_h} (\zeta -\lambda_h)^{-1} {\bm R}(\zeta) \; {\rm d}\zeta,
$$
which satisfies ${\bm P}_h{\bm S}_h = {\bm S}_h{\bm P}_h = {\bm O}$, where 
${\bm O}$ is the zero matrix. 
Additionally, we have 
$
{\bm T}{\bm P}_h = \lambda_h {\bm P}_h + {\bm D}_h,
$
where ${\bm D}_h$ is referred as the {\it eigennilpotent} operator of ${\bm T}$.
For later use, with $q \in {\bf Z}$ we define:
\begin{align}
&{\bm S}_h^{(0)} = -{\bm P}_h, \quad {\bm S}_h^{(q)} = {\bm S}_h^q \; (q>0), 
\quad  {\bm S}_h^{(q)} = {\bm O} \; (q<0).
\end{align}
Here, we have used the fact that ${\bm U}$ is unitary, and therefore normal, meaning that its corresponding eigennilpotent operator vanishes.
Using the same fact, we will later discuss the form of the eigenvalues of ${\bm U}$ under a perturbation.
For simplicity, we omit the subindex $h$ provided there is no ambiguity.
\subsection{Perturbation on eigenvalue of ${\bm T}$}
We now expand $\lambda_h(\chi)$, the eigenvalues of ${\bm T}(\chi)$. Using the resultant representation, we will later find the form of the perturbed eigenvalues of ${\bm U}$.
We first discuss whether $\lambda_h(\chi)$ is holomorphic on a certain region in the complex plane. To answer this question, we need the following facts from~\cite{Kato_1980}.
\begin{thrm}
Let $X$ be an unitary space and let $\chi_0 \in D_0$ for a certain simply connected
region $D_0 \subset {\bf C}$.
In addition, let a sequence $\{ \chi_n\}$ converges to $\chi_0$, and let ${\bm T}(\chi_n) \; (n=1,2,\ldots)$ be normal.

All eigenvalues and eigenprojections of ${\bm T}(\chi)$, denoted as $\lambda_h(\chi)$ and ${\bm P}_h(\chi)$ respectively, are then holomorphic at $\chi=\chi_0$.
\end{thrm}
\begin{corollary}
Let $\{ {\bm T}(\chi) \}_{\chi}$ be a general family of normal matrices,
 that are holomorphic with respect to $\chi \in {\bf R}$. Then, their eigenvalues $\{ \lambda_h(\chi)\}$ are also
holomorphic with respect to $\chi$ on ${\bf R}$.
\end{corollary}
In our case, ${\bm T}$ and ${\bm U}$ are symmetric and unitary operators, respectively, so satisfy Theorem 5.1 and Corollary 5.1, respectively.
Therefore, $\lambda_h(\chi)$ and $\mu_h(\chi)$ are holomorphic on the real axis. Within a neighborhood of the real axis, they can be expressed in the form
\begin{align}
&\lambda_h(\chi) = \lambda_h + \sum_{n=1}^\infty \chi^n {\lambda}_h^{(n)},\quad
\mu_h(\chi) = \mu_h + \sum_{n=1}^\infty \chi^n {\mu}_h^{(n)}.
\end{align}
Hereafter, the first and second equalities in (5.6) are denoted as (5.6)$_1$ and (5.6)$_2$, respectively.

Our next question is: how do we find and estimate the coefficients of the above expansion?
The estimates of coefficients are also needed for finding the error bounds, as discussed later. 
If $\lambda_h$ is a simple eigenvalue of ${\bm T}$, we can write
\begin{align}
&\lambda_h^{(n)} = \sum_{p=1}^n \frac{(-1)^p}{p} 
 \sum_{\nu_1 + \ldots + \nu_p=n \atop k_1+ \ldots + k_p = p-1}
 {\rm tr} \; {\bm T}^{(\nu_1)} {\bm S}_h^{(k_1)} \ldots {\bm T}^{(\nu_p)} {\bm S}_h^{(k_p)}.
\end{align}
We also have an estimate of the form 
\begin{align}
&|\lambda_h^{(n)} | \leq \varrho_h  r_h^{-n},
\end{align}
where $r_h$ is the convergence radius of (5.6)$_1$, and 
$\varrho_h = \max_{\zeta \in \Gamma_h} |\zeta-\lambda_h|$ with $\Gamma_h$ as defined in Section 5.2. 
To find the convergence radius of (5.6)$_1$, we first define $r^{(h)}(\zeta)>0$ where $\zeta \in {\rm \Gamma}_h$ as a number satisfying
$$
\sum_{n=1}^\infty \bigl( r^{(h)}(\zeta) \bigr)^n \|{\bm T}^{(n)} {\bm R}(\zeta)\| =1.
$$
Then, for $|\chi| < r^{(h)}(\zeta)$, 
$
\sum_{n=1}^\infty |\chi|^n \|{\bm T}^{(n)} {\bm R}(\zeta)\| <1
$ holds.
For such $\chi$, the infinite sum
$$
\|{\bm A}(\chi) {\bm R}(\zeta)\| 
 = \left\| \Bigl(\sum_{n=1}^\infty \chi^n  {\bm T}^{(n)}\Bigr) {\bm R}(\zeta) \right\|
$$
converges to a value less than $1$, meaning that (5.2) and (5.3) also converge there. That is, $r_h =\min_{\zeta \in {\Gamma}_h} r^{(h)}(\zeta)$.
Actually, as ${\bm T}$ is normal, the lower bound of $r_h$ is given by
$\bigl( 2\bar{A}/d_h + \bar{B} \bigr)^{-1}$ for some constants $\bar{A}, \; \bar{B}>0$ and the isolation distance $d_{h}$ of $\lambda_h$.
 The convergence radius is then given by 
\begin{align}
&r_1 = \min \bigl\{ \min_{i,j \in V} r_{ij}, \min_{h=1,2,\ldots ,s} \{ r_h \} \bigr\},
\end{align}
where $r_{ij}$ is the value of $\chi$ satisfying $\tilde{g}_{ij}(\chi) = 1$. For lower $n$'s, (5.7) simplifies to
\begin{align*}
&\hspace{30mm} \lambda^{(1)} = \bigl( {\bm T}^{(1)} \phi, \psi \bigr), \\
&\lambda^{(2)} = \bigl( {\bm T}^{(2)} \phi, \psi \bigr)
 - \sum_j (\lambda_j-\lambda_h)^{-1} \bigl( {\bm T}^{(1)} \phi, \psi_j \bigr)
 \bigl( {\bm T}^{(1)} \phi_j, \psi \bigr).
\end{align*}
Here, $\phi$ is an eigenvector corresponding to $\lambda$, 
$\psi$ is the eigenvector of ${\bm T}^*$ corresponding to $\lambda^*$ (where ${\bm T}^*$ is the adjoint operator of ${\bm T}$),
$\{\lambda_j \}$ are the eigenvalues of ${\bm T}$ with corresponding eigenvectors $\{ \phi_j \}$,
and $\{\psi, \psi_j\}$ is the basis of $X^*$ adjoint to the basis $\{\phi, \phi_j\}$ of $X$.

In general, some eigenvalues may not be simple when $\chi=0$, and may split as $\chi$ grows.
In this case, Eq. (5.6)$_1$ only sums the perturbed eigenvalues, and the expansion of each eigenvalue remains unknown.
To calculate the explicit expansion of each eigenvalue, we apply the {\it reduction process} (for details, see \cite{Kato_1980}).

Algorithm 1 shows the overall process for finding the explicit expansions of all eigenvalues, along with descriptions of each process. Our idea is to construct a `tree of eigenvalues'
(see Figure 1). 

\begin{algorithm}
\DontPrintSemicolon 
\KwIn{Google matrix ${\bm G}$}
\KwOut{Explicit representation of each eigenvalue of ${\bm T}(\chi)$}
Define ${\bm T}$ and ${\bm T}(\chi)$ as in Subsections 3.2 and 3.3. \;
Extract the set $\Phi({\bm T})$ of eigenvalues of ${\bm T}$ that are not simple. \;
 \uIf{ $\Phi({\bm T}) = \phi$}{
    Find the explicit expansion of all eigenvalues of ${\bm T}$ by using the formula (5.6)$_1$;
  }
 \Else{
    $k \gets 0$ \;
    %
    \uIf{ $k=0$ }{
       $\Phi_{(k)} = \Phi$;
     }
    \Else{
     Define $\widetilde{\bm T}_{0,1,\ldots,k}$ and $\Phi_{(k)} \equiv \bigl\{ \lambda \in \sigma \bigl( \widetilde{\bm T}_{0,1,\ldots,k} \bigr) \bigr| m(\lambda ) > 1\bigr\}$. \;       
     }
     \While{ $\Phi_{(k)} \ne \phi$ }{
       For eigenvaues in the $k$-th level that are simple, calculate and store their expansions.\;
       Create the level $(k+1)$, $\widetilde{\bm T}_{0,\ldots,k}$ and $\widetilde{\bm T}_{0,\ldots,k}(\chi)$ to compute the descendant eigenvalues of those in $\Phi_{(k)}$, and for each element $\lambda_{j^{(k)}}^{(k)} \in  \Phi_{(k)}$, connect  $\lambda_{j^{(k)}}^{(k)}$ and its descendant eigenvalues in level $(k+1)$. \;
       $k \gets k+1$ \;
     $\Phi_{(k)} \equiv \bigl\{ \lambda \in \sigma \bigl( \widetilde{\bm T}_{0,1,\ldots,k} \bigr) \bigr| m(\lambda ) > 1\bigr\}$. \;
   }
  Trace all paths from $\lambda_h \in \sigma \bigl( {\bm T} \bigr)$ to the ending eigenvalues, to make the expansion of each one in $\sigma \bigl( {\bm T} \bigr)$.
}
\caption{Logical flow of getting expansions of $\lambda_h(\chi)$}
\label{algo:max}
\end{algorithm}
%
\begin{figure}
  \includegraphics[width=13cm]{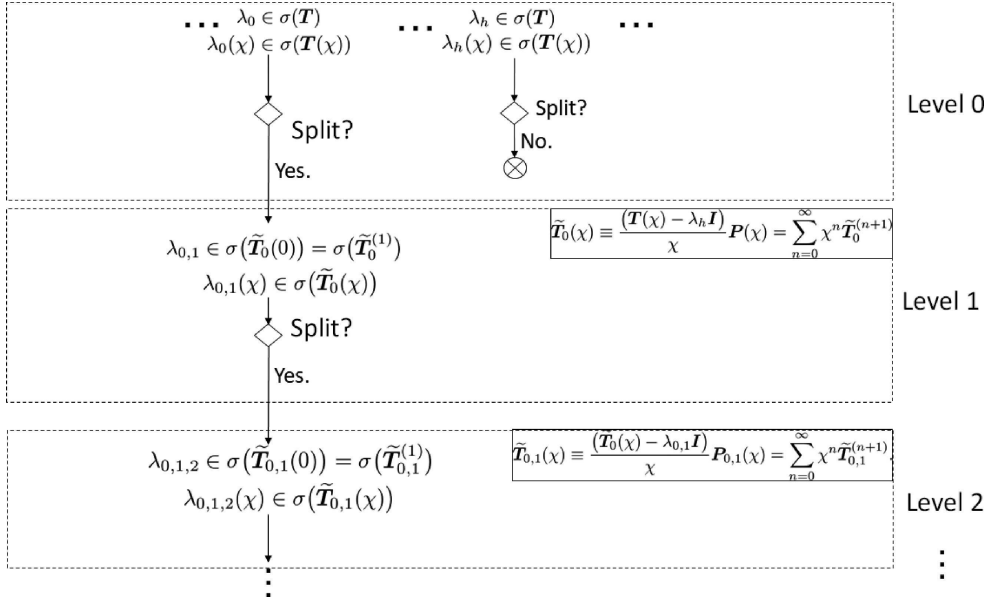}
  \centering
  \caption{Reduction process and construction of the eigenvalue tree}
\end{figure}
The top level contains the eigenvalues of ${\bm T}$, which may include simple and non-simple types.
The edges of the non-simple eigenvalues are extended to the descendant eigenvalues in the next level, found through the reduction process.
On the other hand, the simple eigenvalues have no further edges. We now derive the descendant eigenvalues from the non-simple eigenvalues in the former level. 

For a certain non-simple eigenvalue $\lambda_h$ of ${\bm T}$, we can expand a perturbed eigenprojection of the form (5.3). We thus obtain
\begin{align*}
&\Bigl( {\bm T} (\chi) - \lambda_h {\bm I} \Bigr) {\bm P}_h(\chi) 
 = \sum_{n=1}^\infty \chi^n \widetilde{\bm T}_0^{(n)},
\end{align*}
where 
\begin{align*}
&\widetilde{\bm T}_0^{(n)} 
 = -\frac{1}{2\pi {\rm i}} \sum_{\nu_1 + \nu_2 + \ldots + \nu_p=n \atop \nu_j \geq 1}
%
  \int_{\Gamma_h} {\bm R}(\zeta) {\bm T}^{(\nu_1)} \ldots  {\bm T}^{(\nu_p)}
 {\bm R}(\zeta) (\zeta-\lambda_h) \; {\rm d}\zeta \\
&\hspace{7mm} 
 = - \sum_{p=1}^n (-1)^p
    \sum_{\nu_1 + \nu_2 + \ldots + \nu_p=n \atop k_1+\ldots +k_{p+1} = p-1,
    \nu_j \geq 1, k_j \geq 0}
%
     {\bm S}_h^{(k_1)} {\bm T}^{(\nu_1)} \ldots
      {\bm S}_h^{(k_p)} {\bm T}^{(\nu_p)} {\bm S}_h^{(k_{p+1})}.
\end{align*}
Using this expression, we introduce
$$
\widetilde{\bm T}_0(\chi) \equiv 
\frac{ \bigl( {\bm T}(\chi) - \lambda_h {\bm I} \bigr)}{\chi} {\bm P}_h(\chi)
 = \sum_{n=0}^\infty \chi^n \widetilde{\bm T}_0^{(n+1)},
$$
As ${\bm T}$ is symmetric, $\lambda_h$ is semisimple, so the terms are simplified.
For instance, the spectral decomposition of $\widetilde{\bm T}_0^{(1)}$ in $R({\bm P}_h)$ is given as
\begin{align*}
&\widetilde{\bm T}_0^{(1)} = {\bm P}_h {\bm T}^{(1)} {\bm P}_h
  = \sum_{j^{(1)} } \lambda_{0 ,j^{(1)} } {\bm P}_{0, j^{(1)}}.
\end{align*}
Here, $\{ \lambda_{0,j^{(1)}} \}_{j^{(1)}} $ are the 
eigenvalues of $\widetilde{\bm T}_0^{(1)}$  in $R({\bm P}_h)$.
We call these eigenvalues the descendant eigenvalues of $\lambda_0$ in level-1.
Note that $\widetilde{\bm T}_0^{(1)} = \widetilde{\bm T}_0(0) \equiv \widetilde{\bm T}_0$.
Actually, if we set $\lambda_h$ as the top node of the tree, the descendant eigenvalues $\{ \lambda_{0, j^{(1)}} \}_{j^{(1)}} $ are clearly derived from $\lambda_h$.

At the beginning of the process, we initialize a set of non-simple eigenvalues of ${\bm T}$ as $\Phi_{(0)}$. 
 If $\Phi_{(0)}=\phi$, no further processing is required, and each eigenvalue of ${\bm T}$ is explicitly expanded by Eq. (5.6)$_1$.

Otherwise, we find the descendant eigenvalues of each element in $\Phi_{(0)}$, and initialize a set $\Phi_{(1)}$ of non-simple eigenvalues in level-1 (line 7 in Algorithm 1).
If $\Phi_{(1)} = \phi$, each eigenvalue in level-1 is explicitly expanded by Eq. (5.6)$_1$.
The eigenvalues $\lambda_{0, j^{(1)}} (\chi)$ are then explicitly expanded as 
$$
\lambda_{0, j^{(1)}} (\chi) = \lambda_{0, j^{(1)}}
  + \sum_{n=1}^\infty \chi^n \lambda_{0, j^{(1)}}^{(n)}, 
$$
and the corresponding eigenvalue of ${\bm T}(\chi)$ are expressed as
\begin{align*}
&\lambda_h(\chi) = \lambda_h + \chi \lambda_{0,j^{(1)}}(\chi)\\
&\hspace{10mm}
=\lambda_h + \chi \lambda_{0,j^{(1)}}
  + \sum_{n=1}^\infty \chi^{n+1} \lambda_{0,j^{(1)}}^{(n)}.
\end{align*}
The $ \lambda_{0,j^{(1)}}^{(n)}$ are given by (5.7), with ${\bm T}^{(\nu)}$ and ${\bm S}_h^{(k)}$ replaced by
$\widetilde{\bm T}_{0}^{(\nu+1)}$ and ${\bm S}_1^{(k)}$ (defined later), respectively. 

If $\Phi_{(1)} \ne \phi$, the reduction process is reiterated to find the descendant eigenvalues of those in $\Phi_{(1)}$.

We now explain our procedure in detail, denoting $\lambda_{0,,j^{(1)} }$ by $\lambda_{0,1}$ and ${\bm P}_h$ by ${\bm P}_0$ for simplicity.
Let $\lambda_{0,1} \in \Phi_{(1)}$ be the eigenvalue of $\widetilde{\bm T}_0$, and the descendant eigenvalue of a certain $\lambda_0$.
We also take the corresponding perturbed eigenvalue $\lambda_{0,1}(\chi)$ as the eigenvalue of $\widetilde{\bm T}_0^{(1)}(\chi)$.
The eigenprojection ${\bm P}_{0,1}$ corresponding to this $\lambda_{0,1}$ in $R({\bm P}_0)$ is computed as 
\begin{align*}
&{\bm P}_{0,1} = \int_{\Gamma_{0,1}} {\bm R}_{{0,1}} (\zeta) \; {\rm d}\zeta,
\end{align*}
where 
${\bm R}_{{0,1}} (\zeta) = \bigl(  \widetilde{\bm T}_0 - \zeta {\bm I}\bigr)^{-1}$, and 
$\Gamma_{0,1}$ is a closed curve isolating $\lambda_{0,1}$ from all other eigenvalues of $\widetilde{\bm T}_{0}$.
We thus define 
$\widetilde{\bm T}_{0,1} = {\bm P}_{0,1}  \widetilde{\bm T}_0^{(2)} {\bm P}_{0,1}$,
and
\begin{align}
&{\bm P}_{0,1}^{(n)} = 
  -\sum_{p=1}^n (-1)^p 
    \sum_{\nu_1+ \ldots +\nu_p=n \atop k_1 + \ldots +k_{p+1}=p, \nu_j \geq 1, k_j \geq 0} 
%
    {\bm S}_1^{(k_1)} \widetilde{\bm T}_0^{(\nu_1+1)} \ldots 
     {\bm S}_1^{(k_p)}  \widetilde{\bm T}_0^{(\nu_p+1)}  {\bm S}_1^{(k_{p+1})}, \notag \\
&\widetilde{\bm T}_{0,1}^{(n)} = -\sum_{p=1}^n (-1)^p
\sum_{\nu_1+\ldots +\nu_p = n \atop k_1+\ldots + k_{p+1}=p-1, \; \nu_j \geq 1, \; k_j \geq 0} 
%
{\bm S}_1^{(k_1)} \widetilde{\bm T}_0^{(\nu_1+1)} \ldots
{\bm S}_1^{(k_p)} \widetilde{\bm T}_0^{(\nu_p+1)} {\bm S}_1^{(k_{p+1})}.
\end{align}
Here,
\begin{align*}
&{\bm S}_1^{(k)} = 
 \left\{
   \begin{array}{l}
    \displaystyle
       \Bigl[ {\bm S}_1^\prime - \frac{1}{\lambda_{0,1}} \bigl( {\bm I}-{\bm P}_0 \bigr) \Bigr]^k \quad (k \geq 1),\\[12pt]
   \displaystyle -{\bm P}_{0,1} \quad (k=0)
   \end{array}
 \right.
\end{align*}
with 
$$
{\bm S}_1^\prime \equiv -\sum_{\lambda_{0,1}^\prime \ne \lambda_{0,1}}
 \frac{ {\bm P}_{\lambda_{0,1}^\prime}}{|\lambda_{0,1} - \lambda_{0,1}^\prime|}.
$$
Above we assumed that $\lambda_{0,1} \ne 0$, and ${\bm P}_{\lambda_{0,1}^\prime}$ is an eigenprojection corresponding to $\lambda_{0,1}^\prime$.
The summation runs over all eigenvalues of $\widetilde{\bm T}_{0}$ except $\lambda_{0,1}$. We then define
\begin{align*}
&\hspace{10mm}
 {\bm P}_{0,1}(\chi) =
  {\bm P}_{0,1} + \chi {\bm P}_{0,1}^{(1)} + \chi^2 {\bm P}_{0,1}^{(2)} + \ldots, \\[5pt]
&\widetilde{\bm T}_{0,1} (\chi) \equiv 
 \frac{ \bigl( \widetilde{\bm T}_0(\chi) - \lambda_{0,1} {\bm I}\bigr) }{ \chi} {\bm P}_{0,1}(\chi)
 = \sum_{n=0}^\infty \chi^n \widetilde{\bm T}_{0,1}^{(n+1)}.
\end{align*}
From (5.10), it is seen that
 $\widetilde{\bm T}_{0,1} \equiv \widetilde{\bm T}_{0,1} (0) =  \widetilde{\bm T}_{0,1}^{(1)} = {\bm P}_{0,1} \widetilde{\bm T}_{0}^{(2)} {\bm P}_{0,1}.$
Now, the eigenvalues of $\widetilde{\bm T}_{0,1}$ and $\widetilde{\bm T}_{0,1}(\chi)$ (denoted as $\lambda_{0,1,2}$ and $\lambda_{0,1,2}(\chi)$, respectively)
are the descendant eigenvalues of $\lambda_{0,1}$ in level-2. This procedure is applied to all eigenvalues in $\Phi_{(1)}$ in level-1.
Now, if the eigenvalue $\lambda_{0,1,2}$ of $\widetilde{\bm T}_{0,1}(0)=\widetilde{\bm T}_{0,1}$ is simple, its perturbed value is represented as
$$
\lambda_{0,1,2} (\chi) = \sum_{n=0}^\infty \chi^n \lambda_{0,1,2}^{(n)}.
$$
The $ \lambda_{0,1,2}^{(n)}$ are given by (5.7), replacing ${\bm T}^{(\nu)}$ and ${\bm S}_h^{(k)}$ by $\widetilde{\bm T}_{0,1}^{(\nu+1)}$ and ${\bm S}_2^{(k)}$(defined later), respectively. 
The corresponding eigenvalue of ${\bm T}(\chi)$ is represented as
\begin{align}
&\lambda (\chi) = \lambda_0 + \chi \lambda_{0,1} + \chi^2 \sum_{n=0}^\infty \chi^n \lambda_{0,1,2}^{(n)}.
\end{align}
If all of these eigenvalues are simple, they can be determined by Eq.~(5.6)$_1$, and the procedure terminates.
Otherwise, we should recheck whether all eigenvalues in level-2 are simple. For this purpose, we initialize a set $\Phi_{(2)}$ of non-simple eigenvalues in level-2, and discuss the case
$\Phi_{(2)} \ne \phi$.
We first introduce
\begin{align*}
&{\bm P}_{0,1,2} = \int_{\Gamma_{0,1,2}} {\bm R}_{{0,1,2}} (\zeta) \; {\rm d}\zeta,
\end{align*}
where 
${\bm R}_{{0,1,2}} (\zeta) = \bigl(  \widetilde{\bm T}_{0,1}- \zeta {\bm I}\bigr)^{-1}$, and 
$\Gamma_{0,1,2}$ is a closed curve that isolates $\lambda_{0,1,2}$ from all other eigenvalues of $\widetilde{\bm T}_{0,1}$.
We also define:
\begin{align*}
&{\bm P}_{0,1,2}^{(n)} = -\sum_{p=1}^n (-1)^p
\sum_{\nu_1+\ldots +\nu_p = n \atop k_1+\ldots + k_{p+1}=p, \; \nu_j \geq 1, \; k_j \geq 0} 
%
 {\bm S}_2^{(k_1)} \widetilde{\bm T}_{0,1}^{(\nu_1+1)} \ldots
  {\bm S}_2^{(k_p)} \widetilde{\bm T}_{0,1}^{(\nu_p+1)}
   {\bm S}_2^{(k_{p+1})},\\
&\widetilde{\bm T}_{0,1,2}^{(n)} = -\sum_{p=1}^n (-1)^p
\sum_{\nu_1+\ldots +\nu_p = n \atop k_1+\ldots + k_{p+1}=p-1, \; \nu_j \geq 1, \; k_j \geq 0} 
%
 {\bm S}_2^{(k_1)} \widetilde{\bm T}_{0,1}^{(\nu_1+1)} \ldots
  {\bm S}_2^{(k_p)} \widetilde{\bm T}_{0,1}^{(\nu_p+1)}
   {\bm S}_2^{(k_{p+1})},
\end{align*}
where
\begin{align*}
&{\bm S}_2^{(k)} = 
 \left\{
   \begin{array}{l}
    \displaystyle
       \Bigl[ {\bm S}_2^\prime - \frac{1}{\lambda_{0,1,2}} \bigl( {\bm I}-{\bm P}_{0,1} \bigr) \Bigr]^k \quad (k \geq 1),\\[8pt]
   \displaystyle -{\bm P}_{0,1,2} \quad (k=0)
   \end{array}
 \right.
\end{align*}
with 
$$
{\bm S}_2^\prime \equiv -\sum_{\lambda_{0,1,2}^\prime \ne \lambda_{0,1,2}}
 \frac{ {\bm P}_{\lambda_{0,1,2}^\prime}}{|\lambda_{0,1,2} - \lambda_{0,1,2}^\prime|}.
$$
We assume that $\lambda_{0,1,2} \ne 0$, and let ${\bm P}_{\lambda_{0,1,2}^\prime}$ be an eigenprojection corresponding to $\lambda_{0,1,2}^\prime$.
The summation of the right-hand side runs over all eigenvalues of $\widetilde{\bm T}_{0,1}$ except $\lambda_{0,1,2}$. 
We then introduce
\begin{align*}
&\hspace{17mm}
{\bm P}_{0,1,2}(\chi) = {\bm P}_{0,1,2} + \sum_{j=1}^\infty \chi^j {\bm P}_{0,1,2}^{(j)},\\
&\widetilde{\bm T}_{0,1,2} (\chi) \equiv 
 \frac{ \bigl( \widetilde{\bm T}_{0,1}(\chi) - \lambda_{0,1,2} {\bm I} \bigr) }{ \chi} {\bm P}_{0,1,2}(\chi)
 = \sum_{n=0}^\infty \chi^n \widetilde{\bm T}_{0,1,2}^{(n+1)}.
\end{align*}
Again, we note that
 $\widetilde{\bm T}_{0,1,2} \equiv \widetilde{\bm T}_{0,1,2} (0) 
 = \widetilde{\bm T}_{0,1,2}^{(1)} = {\bm P}_{0,1,2} \widetilde{\bm T}_{0,1}^{(2)} {\bm P}_{0,1,2}.$
Now, the eigenvalues of $\widetilde{\bm T}_{0,1,2}$ and $\widetilde{\bm T}_{0,1,2}(\chi)$,
denoted as $\lambda_{0,1,2,3}$ and $\lambda_{0,1,2,3}(\chi)$, respectively, 
are the descendant eigenvalues of $\lambda_{0,1,2}$ in level-3.
This procedure is applied to all eigenvalues in $\Phi_{(2)}$ in level-2.
Repeating this procedure until none of the eigenvalues split, we obtain the exact representations of all eigenvalues.  
Note that the procedure halts within a finite number of iterations (at most level-N iterations). 
Without loss of generality, we have assumed that all eigenvalues (including the descendant ones) do not vanish. If some eigenvalues do vanish, 
the unit of $\chi$ can be adjusted so that all eigenvalues that appear in the iteration prodedure always remain non-vanishing. 

\begin{remark}
While tracing the tree, the convergence radius should be updated as the minimum of $r_1$ computed by {\rm (5.9)} or the convergence radius of the eigenvalue at the bottom of each path. 
Because this procedure expires after finitely many iterations, we can obtain the final convergence radius, and assign it to $r_1$.
In addition, the estimated expression {\rm (5.8)} holds for all $\lambda_h$.
By resetting the constants, we can replace $\varrho_h$ and $r_h$ with constants independent of $h$, hereafter denoted as $\varrho_1$ and $r_1$, respectively. 
\end{remark}
This fact is summarized as a proof at the end of this subsection. 
\begin{lmm}
Assume that {\rm (3.3)} holds. In the above expressions of $\lambda_h(\chi)$, we have $|\lambda_h^{(n)}| \leq \varrho_1 r_1^n \; (n=1,2,\ldots, \; h=1,2,\ldots, s)$ 
with constants $\varrho_1, \; r_1 > 0$ independent of $h$.
\end{lmm}
\begin{proof}
For the eigenvalues of ${\bm T}$, it is sufficient to assume (5.33) (see Lemma 5.4 for proof). It remains to show that {\rm Lemma 5.1} holds for the descendant eigenvalues.
Now, take a simple eigenvalue $\lambda_{0,1,\ldots,L+1}$ of $\widetilde{\bm T}_{0,1,\ldots,L}$ in level-$(L+1)$, and perturb it as $\lambda_{0,1,\ldots,L+1}(\chi)$.
The perturbed eigenvalue is an eigenvalue of $\widetilde{\bm T}_{0,1,\ldots,L}(\chi) = \sum_{n=0}^\infty \chi^n \widetilde{\bm T}_{0,1,\ldots,L}^{(n+1)}$.
The corresponding eigenvalue of ${\bm T}$ is then denoted as 
\begin{align*}
&\lambda(\chi) = \lambda_h + \chi \l + \lambda_{0,1} + \ldots +\chi^L \lambda_{0,1,\ldots,L}
+\chi^{L+1} \sum_{n=0}^\infty \chi^n \lambda_{0,1,\ldots,L+1}^{(n)}.
\end{align*}
It is sufficient to show that $\widetilde{\bm T}_{0,1,\ldots,L}^{(n)}$ are estimated as 
\begin{align}
&\|\widetilde{\bm T}_{0,1,\ldots,L}^{(n)}\| \leq \widetilde{A}_L \widetilde{B}_L^{n-1} \; n=1,2,\ldots.
\end{align}
We show this by induction. First consider the case $L=0$. As ${\bm T}$ is normal, we have 
$\|{\bm S}_h\| \leq 1/\min_{\lambda_k \ne \lambda_h} |\lambda_k - \lambda_h|.$
Using this, we can write 
\begin{align}
&\|\widetilde{\bm T}_0^{(n)}\| 
 \leq \sum_{p=1}^n 
  \Bigl| \frac{1}{ \min_{\lambda_k \ne \lambda_h} |\lambda_k - \lambda_h| } \Bigr|^{p-1}
   \sum_{\nu_1+\ldots +\nu_p=n \atop k_1 + \ldots + k_{p+1} = p-1} A_2^p B_2^n \notag \\
&\hspace{10mm}
 = B_2^n \sum_{p=1}^n \Bigl| \frac{1}{ \min_{\lambda_k \ne \lambda_h} |\lambda_k - \lambda_h| } \Bigr|^{p-1}
   A_2^p \;_{n-1} C_{n-p} \;_{2p-1} C_{p-1}  \notag  \\
&\hspace{10mm}
 \leq 2^{2n-1} B_2^n C_h^{-1} 
  \Biggl[ C_h A_2 \Bigl( 1 + C_hA_2 \Bigr)^{n-1} \Biggr] ,
\end{align}
where $C_h \equiv \Bigl| \frac{1}{ \min_{\lambda_k \ne \lambda_h} |\lambda_k - \lambda_h| } \Bigr|$.
Thus, (5.12) holds for $L=0$. 
Next, assume that (5.12) holds for $L=(l-1)$. Then, for $L=l$, 
\begin{align*}
&\widetilde{\bm T}_{0,1,\ldots,l}^{(n)} = 
 -\sum_{p=1}^n (-1)^p
   \sum_{\nu_1+\ldots +\nu_p=n \atop k_1 + \ldots + k_{p+1} = p-1,\nu_j \geq 1,k_j \geq 0} 
%
    {\bm S}_l^{(k_1)}  \widetilde{\bm T}_{0,1,\ldots, l-1}^{(\nu_1+1)} \ldots 
    {\bm S}_l^{(k_p)}  \widetilde{\bm T}_{0,1,\ldots, l-1}^{(\nu_p+1)}  {\bm S}_l^{(k_{p+1})}. 
\end{align*}
Assuming (5.12), we have $\| \widetilde{\bm T}_{0,1,\ldots,l-1}^{(n)} \| \leq
 \widetilde{A}_{l-1} \widetilde{B}_{l-1}^{n-1}$.
Recall also that
\begin{align*}
&{\bm S}_{l}^{(k)} = 
 \left\{
   \begin{array}{l}
    \displaystyle
       \Bigl[ {\bm S}_l^\prime - \frac{1}{\lambda_{0,1,\ldots,l}} \bigl( {\bm I}-{\bm P}_{0,1,\ldots,l} \bigr) \Bigr]^k \quad (k \geq 1),\\[12pt]
   \displaystyle -{\bm P}_{0,1\ldots,l} \quad (k=0)
   \end{array}
 \right.
\end{align*}
with 
\begin{align*}
&\hspace{20mm}
 {\bm S}_{l}^\prime \equiv -\sum_{\lambda_{0,1,\ldots,l}^\prime \ne \lambda_{0,1,\ldots,l}}
 \frac{ {\bm P}_{\lambda_{0,1,\ldots,l}^\prime}}{\min|\lambda_{0,1,\ldots,l} - \lambda_{0,1,\ldots,l}^\prime|}, \\
&\hspace{10mm}
{\bm P}_{0,1,\ldots,l} = \int_{\Gamma_{0,1,\ldots,l}} {\bm R}_{0,1,\ldots,l}(\zeta) \; {\rm d}\zeta, \quad
%
{\bm R}_{0,1,\ldots,l}(\zeta) = \Bigl( \widetilde{\bm T}_{0,1,\ldots, l-1}^{(1)} -\zeta {\bm I} \Bigr)^{-1},
\end{align*}
where $\Gamma_{0,1,\ldots,l}$ encloses the eigenvalue $\lambda_{0,1,\ldots,l}$. 
As $\widetilde{\bm T}_{0,1,\ldots,l}^{(1)}$ is normal, we have $\|{\bm P}_{0,1,\ldots,l}\|$ and $\| {\bm P}_{\lambda_{0,1,\ldots,l}^\prime} \| = 1$. We also observe that 
\begin{align*}
&\|{\bm S}_l^{(k)}\| \leq 
  \max \Biggl\{ 1, \Bigl( \frac{1}{\min |\lambda_{0,1,\ldots,l} - \lambda_{0,1,\ldots,l}^\prime| } 
       + \frac{1}{\lambda_{0,1,\ldots,l} }
          \Bigr)^k \Biggr\} 
%
  \equiv 
  \max \bigl\{ 1, C_l^k \bigr\}.
\end{align*}
We thus arrive at
\begin{align*}
&\| \widetilde{\bm T}_{0,1,\ldots,l}^{(n)}\| \leq \sum_{p=1}^n
   \sum_{\nu_1+\ldots +\nu_p=n \atop k_1 + \ldots + k_{p+1} = p-1,\nu_j \geq 1,k_j \geq 0}
%
   \max\{ 1, C_l^{p-1} \} \widetilde{A}_{l-1}^{p} \widetilde{B}_{l-1}^n \\
&\hspace{13mm}
  = \widetilde{B}_{l-1}^n \sum_{p=1}^n 
     \max\{ 1, C_l^{p-1} \} \widetilde{A}_{l-1}^p \;_{n-1}C_{n-p} \;_{2p-1}C_{p-1},
\end{align*}
where $\widetilde{A}_{l-1}$ and $\widetilde{B}_{l-1}$ are some positive constants.
The desired estimate $\| \widetilde{\bm T}_{0,1,\ldots,l}^{(n)} \| \leq \widetilde{A}_l \widetilde{B}_l^{n-1}$ is then deduced as described for (5.13).
\end{proof}

\subsection{Perturbation on ${\bm U}$}
This subsection discusses the form of ${\bm U}$ under a perturbation. We first consider the expansion of 
%
$\ket{\psi_j (\chi)} 
  = \ket{j}_1 \otimes \sum_{k=1}^N \sqrt{g_{jk}(\chi)} \ket{k}_2$
 with respect to $\chi$. For this purpose, we find the expansion of $\sqrt{g_{jk}(\chi)}$.
Now, $\sqrt{g_{jk}(\chi)} = \sqrt{g_{jk} (1+\breve{g}_{jk}(\chi))}$, where 
$\breve{g}_{jk}(\chi) \equiv (g_{jk}(\chi) - g_{jk})/{g_{jk}}$.
The coefficient of $\sqrt{ 1+\breve{g}_{jk}(\chi)} $ of $\chi^n$, denoted as $a_{(j,k)}^{(n)}$, is represented as
\begin{align*}
&a_{(j,k)}^{(n)} = \frac{1}{n!}  \Bigl( \frac{{\rm d}}{{\rm d} \chi} \Bigr)^n
 \Bigl[ \sqrt{ 1+\breve{g}_{jk}(\chi)} \Bigr]
  \Bigr|_{\chi=0}.
\end{align*}
Now, 
\begin{align*}
& \Bigl( \frac{{\rm d}}{{\rm d} \chi} \Bigr)^n
    \Bigl[ \sqrt{ 1+\breve{g}_{jk}(\chi)} \Bigr]
     \Bigr|_{\chi=0}
%
 = \sum_{r=1}^n \;_\frac12 C_r
     \sum_{p_1+\ldots p_r=n \atop p_j \geq 1}
      \frac{n !}{p_1 ! \ldots p_r !} \prod_{i=1}^r 
       \Bigl( \frac{p_i! g_{jk}^{(p_i)}}{g_{jk}}\Bigr) \\
&\hspace{40mm}
 =  n! \sum_{r=1}^n \;_\frac12 C_r \sum_{p_1+\ldots p_r=n \atop p_j \geq 1}
      \prod_{i=1}^r  \frac{ g_{jk}^{(p_i)} }{g_{jk}}.
\end{align*}
Therefore, we have 
\begin{align*}
&a_{(j,k)}^{(n)} = \sum_{r=1}^n \;_\frac12 C_r \sum_{p_1+\ldots p_r=n \atop p_j \geq 1}
          \prod_{i=1}^r  \frac{ g_{jk}^{(p_i)} }{g_{jk}},
\end{align*}
which gives
\begin{align}
&\ket{\psi_j(\chi)} 
%
 = \ket{\psi_j} + \sum_{n=1}^\infty \ket{\psi_j^{(n)}} \chi^n.
\end{align}
The convergence radius of (5.14) is lower-bounded by $r_1$. 
${\bm B}(\chi)$ is then expanded as 
\begin{align}
&{\bm B}(\chi) = \sum_{j=1}^N \ket{\psi_j(\chi)}\bra{\psi_j(\chi)}
 = {\bm B} + \sum_{n=1}^\infty {\bm B}^{(n)} \chi^n, 
\end{align}
where 
$$
{\bm B}^{(n)}  
= \sum_{j=1}^N \sum_{l=0}^n \ket{\psi_j^{(l)}} \bra{\psi_j^{(n-l)}}.
$$
For simplicity, we have used the notation $\ket{\psi_j^{(0)}} = \ket{\psi_j}$. Expanded as (5.14) and (5.15), the operator ${\bm U}$ under a perturbation is represented as 
\begin{align}
&{\bm U}(\chi) = {\bm S}_w \Bigl( 2 {\bm B}(\chi) - {\bm I} \Bigr)
 = {\bm U} +  \sum_{n=1}^\infty {\bm U}^{(n)} \chi^n,
\end{align}
where 
\begin{align}
&{\bm U}^{(n)} = 2 {\bm S}_w  {\bm B}^{(n)}
 = 2{\bm S}_w \sum_{j=1}^N \sum_{l=0}^n 
  \ket{\psi_j^{(l)}} \bra{\psi_j^{(n-l)}}.
\end{align}
The infinite sum (5.16) converges as far as $\ket{\psi(\chi)}$, so their convergence radii are also $r_1.$
We now define the eigenprojection of ${\bm U}$ for each eigenvalue of ${\bm U}$.

Note that ${\bm U}$ and ${\bm U}(\chi)$ are unitary, and consequently semisimple. The resolvent of ${\bm U}$ is defined as 
$
\hat{\bm R}(\zeta) \equiv ({\bm U} -\zeta {\bm I} )^{-1},
$
and the eigenprojection corresponding to eigenvalue $\mu_h$ (recall that $\mu_h$ can be $\pm 1$) is 
$$
\hat{\bm P}_h \equiv
  -\frac{1}{2\pi {\rm i}} \int_{ \hat{\Gamma}_h } \hat{\bm R}(\zeta) \; {\rm d}\zeta.
$$
Here, $\hat{\Gamma}_h$ is an arbitrary convex loop in the complex plain that isolates $\mu_h$ from all other eigenvalues of ${\bm U}$. 
We now represent the resolvent and eigenprojection under a small perturbation. 
We denote the resolvent of ${\bm U}(\chi)$ by $\hat{\bm R}(\zeta,\chi) \equiv \bigl( {\bm U}(\chi) - \zeta {\bm I}\bigr)^{-1}$, where 
$\hat{\bm R}(\zeta,\chi)$ is defined for all $\zeta \notin \sigma \bigl({\bm U}(\chi)\bigr)$. We know that for such a $\zeta$, 
 $\hat{\bm R}(\zeta,\chi)$ is holomorphic with respect to both $\zeta$ and $\chi$, and is hence expanded as follows: 
\begin{align}
&\hat{\bm R}(\zeta,\chi) 
 =\hat{\bm R}(\zeta) \bigl[ {\bm I} + \hat{\bm A}(\chi) \hat{\bm R}(\zeta) \bigr]^{-1} 
%
 = \hat{\bm R}(\zeta) + \sum_{l=1}^\infty \chi^l \hat{\bm R}^{(l)}(\zeta),
\end{align}
where $\hat{\bm A}(\chi) = {\bm U}(\chi) - {\bm U}$, and 
\begin{align*}
&\hat{\bm R}^{(l)}(\zeta) \equiv
  \sum_{\nu_1+\ldots + \nu_p=l \atop \nu_j \geq 1}
 (-1)^p \hat{\bm R}(\zeta) {\bm U}^{(\nu_1)}
            \hat{\bm R}(\zeta) {\bm U}^{(\nu_2)} \ldots 
             \hat{\bm R}(\zeta) {\bm U}^{(\nu_p)}  {\bm R}(\zeta). 
\end{align*}
Integrating the infinite series (5.18) over $\hat{\Gamma}_h$, we have
\begin{align}
&\hat{\bm P}_h(\chi) = -\frac{1}{2\pi {\rm i}} \int_{ \hat{\Gamma}_h}
   \hat{\bm R}(\zeta,\chi) \; {\rm d}\zeta
    = \hat{\bm P}_h + \sum_{l=1}^\infty \chi^l \hat{\bm P}_h^{(l)}.
\end{align}
Equation (5.19) satisfies
$\hat{\bm P}_h(0) = \hat{\bm P}_h
 = -\frac{1}{2\pi {\rm i}} \int_{ \hat{\Gamma}_h} \hat{\bm R}(\zeta) \; {\rm d}\zeta.
$
In (5.19), $\hat{\bm P}_h^{(l)}$ is given by
\begin{align}
&\hat{\bm P}_h^{(l)}  
 = - \frac{1}{2\pi {\rm i}}  \sum_{\nu_1+\ldots + \nu_p=l \atop \nu_k \geq 1}
   (-1)^p \int_{ \hat{\rm \Gamma}_h}
    \hat{\bm R}(\zeta)  {\bm U}^{(\nu_1)}
%
     \hat{\bm R}(\zeta)  {\bm U}^{(\nu_2)} \ldots
      \hat{\bm R}(\zeta)  {\bm U}^{(\nu_p)}
       \hat{\bm R}(\zeta)  \; {\rm d}\zeta.
\end{align}
We also define
$$
\hat{\bm S}_h =  -\frac{1}{2\pi {\rm i}}
 \int_{ \hat{\Gamma}_h } (\zeta -\mu_h)^{-1} 
  \hat{\bm R}(\zeta) \; {\rm d}\zeta,
$$
which satisfies $\hat{\bm P}_h \hat{\bm S}_h= \hat{\bm S}_h\hat{\bm P}_h= {\bm O}$.
Additionally, we have 
$
{\bm U} \hat{\bm P}_h = \mu_h \hat{\bm P}_h .
$
For later use, when $q \in {\bf Z}$ we define:
\begin{align}
&\hat{\bm S}_h^{(0)} = -\hat{\bm P}_h, \quad \hat{\bm S}_h^{(q)} = \hat{\bm S}_h^q \; (q>0), 
\quad  \hat{\bm S}_h^{(q)} = {\bm O} \; (q<0).
\end{align}
Here, we have used the fact that ${\bm U}$ is unitary, and therefore normal.
Later, these expressions will be used for deriving the transformation function of $\{ \ket{\mu} \}$.
\subsection{Perturbation on the eigenvalues of ${\bm U}$}
We now represent the eigenvalues of ${\bm U}(\chi)$. 
%
%

The eigenvalues of ${\bm U}$ (other than those of $-{\bm S}_w$) are given by
$$
\mu = \lambda \pm {\rm i} \sqrt{ 1-  \lambda^2}.
$$
%
Accordingly, the eigenvalues of ${\bm U}(\chi)$ are $\mu(\chi) = \lambda(\chi) + {\rm i} \Bigl( 1- \bigl( \lambda(\chi) \bigr)^2 \Bigr)^\frac12$
for $\chi$ sufficiently close to the real axis, where $\lambda(\chi)$'s are the eigenvalues of ${\bm T}(\chi)$.
Note that $|\chi|$ is sufficiently small that $|\lambda(\chi)|<1$.
Therefore, denoting by $r_2$ the value $|\lambda(\chi)|=1$ for $|\chi|=r_2$, the convergence radius of (5.6) is $r_0 = \min\{ r_1, r_2\}.$
As far as $|\lambda(\chi) |<1$ is satisfied, the branches of  
$\Bigl( 1- \bigl( \lambda(\chi) \bigr)^2 \Bigr)^\frac12$ are holomorphic with respect to $\chi$.
Therefore, expanding this $\mu(\chi)$ as a series of $\chi$, we get
$$
\mu(\chi) = \mu + \sum_{n=1}^\infty \chi^n \mu^{(n)}.
$$
To show this, we first expand $\sqrt{ 1- \bigl( \lambda(\chi) \bigr)^2},$ and recall 
$$
\mu^{(n)} = \frac{1}{n!}\Bigl( \frac{{\rm d}}{{\rm d}\chi} \Bigr)^n \mu(\chi) \Biggr|_{\chi=0}
\; (n=1,2,\ldots). 
$$
We now discuss 
\begin{align}
&
 \frac{1}{n!}
 \Bigl( \frac{{\rm d}}{{\rm d}\chi} \Bigr)^n 
 \Bigl( \sqrt{1-(\lambda(\chi))^2} \Bigr)
 \Biggr|_{\chi=0}. 
\end{align} 
Setting $f(z)=\sqrt{1-z^2}$, we have 
\begin{align}
& \Bigl( \frac{{\rm d}}{{\rm d}\chi} \Bigr)^n \Bigl( f(\lambda(\chi)) \Bigr) \Biggr|_{\chi=0}
 = \sum_{r=1}^n \frac{D^r f(\lambda(\chi))) }{r!} 
 %
  \sum_{p_1+\ldots +p_r=n \atop p_i\geq 1}
   \frac{n!}{ p_1! \ldots p_r!} 
    \Bigl( D^{p_1} \lambda(\chi) \Bigr) \ldots 
     \Bigl( D^{p_r} \lambda(\chi) \Bigr)\Biggr|_{\chi=0} \notag \\
&\hspace{33mm}
 =  n! \sum_{r=1}^n \frac{D^r f(\lambda(\chi))) }{r!} \Bigr|_{\chi=0}
 \sum_{p_1+\ldots +p_r=n \atop p_i \geq 1}
\prod_{j=1}^r \lambda^{(p_j)},
\end{align}
where we have used 
$D^{p_j} \lambda(\chi) \bigr|_{\chi=0} = p_j ! \lambda^{(p_j)}.$
To express the term $D^r f$ in (5.23) above, we set $h(w)=\sqrt{1-w}$ and $l(t)=t^2$, thus obtaining
\begin{align*}
&
\Bigl( \frac{{\rm d}}{{\rm d}t} \Bigr)^r h(l(t))
 = \sum_{r^\prime=1}^r \frac{D^{r^\prime} h(l(t))) }{r^\prime!}
%
 \sum_{p_1^\prime+\ldots +p_{r^\prime}^\prime=r \atop p_i^\prime \geq 1}
 \frac{r!}{ p_1^\prime ! \ldots p_{r^\prime}^\prime !} 
 \Bigl( D^{p_1^\prime} l\Bigr) \ldots 
 \Bigl( D^{p_{r^\prime}^\prime} l\Bigr).
\end{align*}
We also note that $p_i^\prime \in \{1,2\} \; (i=1,2,\ldots,r^\prime)$ and 
\begin{align*}
&D^{p_i^\prime} l = 
 \left\{
 \begin{array}{l}
 2t \quad (p_i^\prime=1), \\[7pt]
 2 \quad (p_i^\prime=2).
 \end{array}
 \right.
\end{align*} 
On the other hand, we can write
\begin{align*}
&D^{r^\prime} \Bigl[ \sqrt{1-w} \Bigr] 
 = (-1)^{r^\prime} r^\prime ! \;_{\frac12} C_{r^\prime}  (1-w)^{-\frac{(2r^\prime-1)}{2}} .
\end{align*}
We thus have 
\begin{align}
&\mu^{(n)} = \lambda^{(n)} + {\rm i} 
  \sum_{r=1}^n
 \Biggl[
   \sum_{r^\prime=1}^r
   (-1)^{r^\prime} \;_{\frac12} C_{r^\prime}  (1-\lambda^2)^{-\frac{(2r^\prime-1)}{2}} 
%
  \sum_{p_1^\prime+\ldots +p_{r^\prime}^\prime=r \atop p_i^\prime>0}
   \frac{2^{r^\prime} \lambda^{(2r^\prime - \sum p_i^\prime)} }{ p_1^\prime ! 
    \ldots p_{r^   \prime}^\prime !} 
 \Biggr] 
%
 \sum_{p_1+\ldots +p_r=n \atop p_i \geq 1}
 \prod_{j=1}^r \lambda^{(p_j)},
\end{align}
which gives an explicit representation of $\mu(\chi)$.
\subsection{Perturbation on the eigenvectors of ${\bm U}$}
This subsection considers the perturbed eigenvectors of ${\bm U}$. Note that the space ${\mathcal H} = {\bm C}^N \otimes {\bf C}^N$
has the orthogonal decomposition:
$$
{\mathcal H} = H_d + H_d^{\perp},
$$
where $H_d$ is the space spanned by $\{ \ket{\psi_j} \}_j$
 and $\{ {\bm S}_w \ket{\psi_j} \}_j$ (called {\it the dynamical space} in~\cite{Paparo_2012}),
 on which ${\bm U}$ operates as an non-trivial operator. Meanwhile, on $H_d^\perp$ (which is spanned by vectors orthogonal to $\{ \ket{\psi_j} \}_j$), 
${\bm U}$ acts as $-{\bm S}_w$. 

As ${\bm U}$ is unitary, its eigenvectors (here called eigenstates) normalized to unit length form an orthonormal basis in ${\mathcal H}$.
Especially, the eigenvectors in $H_e$ form an orthonormal basis there.
Because the imposed perturbation preserves the unitarity of ${\bm U}(\chi)$, we can take a set of its eigenvectors $\{ \ket{ \mu (\chi) } \}$ of unit length as an orthogonal basis of $H_e$ under small perturbation. This method will be shown later on.

Similarly, a set of perturbed eigenvectors $\{ \ket{\mu(\chi)} \}$ corresponding to the eigenvalues of $-{\bm S}_w$ forms an orthogonal basis of the space orthogonal to $H_e$
(denoted as $H_e^{\perp}$). 
Now, applying the transformation function, we can write 
\begin{align*}
&\ket{ \mu(\chi) }= {\bm V}(\chi) \ket{ \mu},
\end{align*}
where ${\bm V}(\chi)$ is an oprator satisfying
\begin{align}
\left\{
 \begin{array}{l} 
 \displaystyle 
 \frac{ {\rm d} }{ {\rm d}\chi} {\bm V}(\chi) = {\bm Q}(\chi){\bm V}(\chi), \\[10pt]
  \displaystyle  {\bm V}(0) = {\bm I},
 \end{array}
 \right.
\end{align}
with ${\bm Q}(\chi) = - \sum_{h=1}^s \hat{\bm P}_h(\chi)
 \frac{ {\rm d} }{ {\rm d}\chi} \hat{\bm P}_h(\chi)$. 
As ${\bm Q}(\chi)$ is holomorphic within a certain region of ${\bf C}$, Eq. (5.25) is guaranteed to have a unique holomorphic solution
 ${\bm V}(\chi)$ with an inverse matrix ${\bm V}(\chi)^{-1}$.
Actually, this solution satisfies 
$$
{\bm V}(\chi) {\bm P}(0) {\bm V}(\chi)^{-1} = {\bm P}(\chi) \quad (h=1,2,\ldots,s).
$$
Using the expansion of  $\hat{\bm P}_h(\chi)$ in (5.19), we expand ${\bm Q}(\chi)$ as a series in $\chi$:
\begin{align*}
&{\bm Q}(\chi) = 
 -\sum_{h=1}^s \Bigl\{ \hat{\bm P}_h  
  + \sum_{n^\prime=1}^\infty \chi^{n^\prime} \hat{\bm P}_h^{(n^\prime)} \Bigr\}
%
   \Bigl\{ \hat{\bm P}_h^{(1)} 
    + \sum_{n=1}^\infty (n+1) \chi^n \hat{\bm P}_h^{(n+1)} \Bigr\} \\
&\hspace{8mm}
 \equiv {\bm Q} + \chi {\bm Q}^{(1)} + \chi^2 {\bm Q}^{(2)}  \ldots.
\end{align*}
Comparing the coefficients, we have 
\begin{align}
&\hspace{35mm}
  {\bm Q} = -\sum_{h=1}^s \hat{\bm P}_h \hat{\bm P}_h^{(1)}, \notag \\
&{\bm Q}^{(r)} = -\sum_{h=1}^s
 \Biggl[
   (r+1)\hat{\bm P}_h \hat{\bm P}_h^{(r+1)} 
%
+ \sum_{r_1=1}^{r} (r-r_1+1) \hat{\bm P}_h^{(r_1)} \hat{\bm P}_h^{(r-r_1+1)}
 \Biggr] \notag \\
&\hspace{65mm}
 \; (r=1,2,\ldots ).
%
\end{align}
For instance, 
${\bm Q}^{(1)} = -2 \sum_{h=1}^s \hat{\bm P}_h\hat{\bm P}_h^{(2)}-\sum_{h=1}^s
 \bigl( \hat{\bm P}_h^{(1)}\bigr)^2.$
We now seek the solution to (5.25) as an infinite series of $\chi$:
$$
 {\bm V} =  {\bm V}_0 + \sum_{n=1}^\infty \chi^n  {\bm V}^{(n)},
$$
and formally represent ${\bm V}^{(n)}$ by comparing the coefficients.
We then observe that ${\bm V}^{(n)}$'s are estimated from above with certain quantities in order to estimate the convergence radius.

First, from the initial condition, we have $ {\bm V}_0 = {\bm I}$.
It is then easily seen that  
\begin{align*}
&{\bm V}^{(1)} = {\bm Q}, \;
  {\bm V}^{(2)} = \frac{{\bm Q}^2 + {\bm Q}^{(1)}}{2}, \;
  {\bm V}^{(3)} = \frac{{\bm Q}^3 + {\bm Q} {\bm Q}^{(1)}}{6}.
\end{align*}
Likewise, we observe that 
\begin{align}
&{\bm V}^{(n)}  = \frac{ \sum_{j=0}^{n-1} {\bm Q}^{(j)} {\bm V}^{(n-1-j)}}{n} \quad (n=4,5,6,\ldots).
\end{align}

Using this, we expand $\ket{ \mu(\chi)}$ as
\begin{align*}
&\ket{ \mu(\chi) } = \ket{ \mu} + \sum_{n=1}^\infty \chi^n \ket{ \mu^{(n)}},
\end{align*}
where 
\begin{align}
&\ket{ \mu^{(n)}} = {\bm V}^{(n)} \ket{ \mu}.
\end{align}
When discussing PageRank, it is sufficient to consider the $\ket{\mu} \in H_e.$
Near the real axis, $I_q(\chi)$ can be expanded as follows. First we have 
We have 
\begin{align*}
& N_q^{(\chi)}( i,m|\psi(0) ) = N_q( i,m |\psi(0)) 
 + \sum_{r=1}^\infty \chi^r N_q^{(r)}( i,m |\psi(0))  \quad \forall i,
\end{align*}
where 
\begin{align}
&N_q^{(n)}( i,m |\psi(0))  
%
= \sum_{\mu \in H_e \atop j \in V}
 \sum_{0 \leq r_1+r_2 \leq n \atop r_j \geq 0}
\Bigl[
\mu^{2m} \sum_{r^\prime=1}^{\min\{r_1,2m\}} \;_{2m} C_{r^\prime}
 \mu^{-r^\prime} 
%
 \sum_{p_1+\ldots + p_{r^\prime}=r_1 \atop p_i \geq 1} 
 \prod_{\tau=1}^{r^\prime} \mu^{(p_\tau)}
\Bigr] \notag \\
&
\hspace{40mm}
 \times
 \braket{\mu^{(r_2)} | \psi(0)}
 \braket{ j,i | \mu^{(n-r_1-r_2)} }.
\end{align}
This is derived by observing the coefficients of $\chi^{r_1}$ in the expansion of $\bigl( \mu(\chi) \bigr)^{2m}$:
\begin{align*}
&\mu^{2m} \sum_{r^\prime=1}^{\min\{r_1,2m\}} \;_{2m} C_{r^\prime}
 \mu^{-r^\prime}
 \sum_{p_1+\ldots + p_{r^\prime}=r_1 \atop p_i \geq 1} 
 \prod_{\tau=1}^{r^\prime} \mu^{(p_\tau)}.
\end{align*}
The derivation follows that of $t_{ij}(\chi)$ in Subsection 5.1.  The coefficients $a_1(n;m)$ of $\chi^n$ in the expansion of $(\mu(\chi))^{2m}$ are denoted as
\begin{align*}
&a_1(n;m) = \frac{1}{n!} 
 \Biggl( 
   \frac{{\rm d}}{{\rm d}\chi}
 \Biggr)^n
 \bigl( \mu(\chi) \bigr)^{2m} \bigr|_{\chi=0} 
%
= \frac{\mu^{2m}}{n!} 
 \Biggl( 
   \frac{{\rm d}}{{\rm d}\chi}
 \Biggr)^n
 \bigl( 1 + \tilde{\mu}(\chi) \bigr)^{2m} \bigr|_{\chi=0},
\end{align*}
where $\tilde{\mu}(\chi)  = (\mu(\chi) -\mu)/\mu = \sum_{n=1}^\infty \chi^n \mu^{(n)}/\mu.$
Then, setting $f(z)=(1+z)^{2m}$, and exploiting $D^{p_j} \tilde{\mu}(\chi) \bigr|_{\chi=0} = p_j ! \mu^{(p_j)}/\mu$, we have 
\begin{align}
&\Bigl( \frac{{\rm d}}{{\rm d}\chi} \Bigr)^n \Bigl( f(\tilde{\mu}( \chi )) \Bigr)\Bigr|_{\chi=0}
 = \sum_{r=1}^n \frac{D^r f(\tilde{\mu}(\chi ))) }{r!} 
%
  \sum_{p_1+\ldots +p_r=n \atop p_i \geq 1}
   \frac{n!}{ p_1! \ldots p_r!} 
    \Bigl( D^{p_1} \tilde{\mu} \Bigr) \ldots 
     \Bigl( D^{p_r} \tilde{\mu} \Bigr) \notag \\
&\hspace{34mm}
 =  n! \sum_{r=1}^n \frac{D^r f(\tilde{\mu}( \chi))) }{r! \mu^r} \Bigr|_{\chi=0}
 \sum_{p_1+\ldots +p_r=n \atop p_i>0}
  \prod_{j=1}^r \mu^{(p_j)}.
\end{align}
Because
\begin{align*}
&D^r \Bigl[ \bigl( 1 + z \bigr)^{2m} \Bigr]
= \left\{
  \begin{array}{l}
   \;_{2m} C_r
   (1+z)^{2m-r} 
  \quad (r=0,1,\ldots, 2m), \\[3pt]
  0 \; (r \geq 2m+1).
 \end{array}
 \right.
\end{align*}
the index $r$ in (5.30) actually runs from $1$ to $\min \{ 2m, n\}.$ 
From these considerations, we have
\begin{align}
&a_1(n;m) = \mu^{2m} \sum_{r=1}^{\min \{ 2m, n\}}
 \;_{2m} C_{r}  \mu^{-r} \sum_{p_1+\ldots +p_r=n \atop p_i>0}
\prod_{j=1}^r \mu^{(p_j)},
\end{align}
which yields (5.29).
We now seek the form of $I_q^{(r)}(i,m|\psi(0) ) \; (r=1,2,\ldots).$ By definition, we have 
\begin{align*}
& I_q^{(\chi)}(i,m|\psi(0) ) =
 \Biggl( 
   \sum_{ \mu(\chi) } \bigl( \mu(\chi) \bigr)^{2m} \sum_j \braket{ j, i | \mu(\chi)}
  \braket{ {\mu}(\chi) | {\psi}(0) } 
 \Biggr) \\
&\hspace{50mm}
 \times \Biggl( 
   \sum_{ \mu(\chi) } \bigl( \mu(\chi) \bigr)^{2m} \sum_j \braket{ j, i | \mu(\chi)}
  \braket{ \mu(\chi) | \psi(0) } 
 \Biggr)^* \\
&\hspace{23mm}
 = \Biggl( N_q ( i,m |\psi(0)) 
     + \sum_{n=1}^\infty \chi^n N_q^{(n)} ( i,m |\psi(0))  \Biggr) \notag \\
&\hspace{50mm}
 \times 
  \Biggl( N_q ( i,m |\psi(0))
     + \sum_{n=1}^\infty \chi^n N_q^{(n)}( i,m |\psi(0))
   \Biggr)^* \\
&\hspace{23mm}
 = I_q(i,m|\psi(0) ) + \sum_{n=1}^\infty \chi^n I_q^{(n)}(i,m|\psi(0) )
\end{align*}
By comparing the coefficients, we have 
\begin{align}
&I_q^{(r)} (i,m | \psi(0)) 
 = N_q^{(r)} ( i,m |\psi(0)) N_q^* ( i,m |\psi(0)) \notag \\
&\hspace{40mm}
  + \sum_{i=1}^{r-1} N_q^{(r-i)} ( i,m |\psi(0)) N_q^{(i)*}( i,m |\psi(0)) \notag \\
&\hspace{40mm}
  + N_q( i,m |\psi(0)) N_q^{(r)*} ( i,m |\psi(0)) 
%
 \quad (r=1,2,\ldots).
\end{align}

\subsection{Error bounds}
We now estimate the latter part of {\rm Theorem 4.1}. To this end, we estimate each term $I_q^{(l)}(i,m|\psi(0)) \; (l=1,2,\ldots,)$ in 
the perturbed quantum PageRank. We first prepare  the following lemmas.
\begin{lmm}
Assume {\rm (3.3)} holds. Then, for $l=1,2,\ldots,$ we have 
\begin{align*}
&\| \ket{\psi_j^{(l)}} \| \leq A_1B_1^{l-1} \; (j=1,2,\ldots ,N, \ l=1,2,\ldots),
\end{align*}
where the constants $A_1,\; B_1>0$ depend on $N$ and ${\bm G}$.
\end{lmm}
\begin{proof}
This can be proved using the representation of $\ket{\psi_j(\chi)}$ (Eq. (5.14)  in Subsection 5.4).
As $\ket{j}_1$, $\ket{k}_2$ are unit vectors, it is sufficient to estimate $a_{(j,k)}^{(n)}$. Under assumption (3.3), we have 
\begin{align*}
&\Bigl| 
   \sum_{k=1}^N 
     \Bigl[
      \sum_{r=1}^n g_{jk}^{\frac12-r} \;_\frac12 C_r
       \sum_{p_1 + \ldots + p_r=n \atop p_j \geq 1}
        \Bigl(
          \prod_{i=1}^r g_{jk}^{(p_i)} 
        \Bigr) 
     \Bigr]
  \Bigr|  
%
%
 \leq    \sum_{k=1}^N 
     \Bigl[
      \sum_{r=1}^n g_{jk}^{\frac12-r}  \bigl| \;_\frac12 C_r \bigr|       
       \sum_{p_1 + \ldots + p_r=n \atop p_j \geq 1}
        A_0^r B_0^{n-r}
      \Bigr] \\
&\hspace{63mm}
 \leq \sum_{k=1}^N 
   \sum_{r=1}^n g_{jk}^{\frac12-r} \bigl| \;_\frac12 C_r \bigr|       
    A_0^r B_0^{n-r} \;_{n-1} C_{r-1} \\
&\hspace{63mm}
 = g_{jk}^{\frac12} B_0^n \sum_{k=1}^N \sum_{r=1}^n
     \bigl| \;_\frac12 C_r \bigr|       \bigl( A_0/{B_0 g_{jk}} \bigr) ^r
      \;_{n-1} C_{r-1} \\
\end{align*}
Taking $\delta \in (0,1)$, and defining
$
A_0/{B_0 g_{jk}} = \delta \times A_0/{\delta B_0 g_{jk}} \equiv \delta \varepsilon_0,
$
we have
\begin{align*}
&g_{jk}^{\frac12} B_0^n \sum_{k=1}^N \sum_{r=1}^n
   \bigl| \;_\frac12 C_r \bigr|       \bigl( A_0/{B_0 g_{jk}} \bigr) ^r
    \;_{n-1} C_{r-1} 
%
  \leq
   g_{jk}^{\frac12} B_0^n \sum_{k=1}^N
    \Biggl( \sum_{r=1}^n \bigl| \;_\frac12 C_r \bigr| \delta^r \Biggr)
     \Biggl( \sum_{r=1}^n \varepsilon_0^r \;_{n-1} C_{r-1} \Biggr) \notag \\[5pt]
&\hspace{63mm}
 = g_{jk}^{\frac12} B_0^n N
    \Bigl\{ 
      1-(1-\delta)^\frac12
   \Bigr\}
     \varepsilon_0 \bigl( 1 + \varepsilon_0 \bigr)^{n-1}.
\end{align*}
Now taking $A_1 = N g_{jk}^{\frac12} \Bigl\{ 1-(1-\delta)^\frac12 \Bigr\} \varepsilon_0$
and $B_1= (1+\varepsilon_0)$, we arrive at our statement.
\end{proof}
From the above, we have 
\begin{lmm}
Assume {\rm (3.3)} holds. Then, for $n=1,2,\ldots,$ we have
\begin{align*}
&\| {\bm U}^{(n)} \| \leq A_2B_2^{n-1} \; (n=1,2,\ldots),
\end{align*}
where the constants $A_2,\; B_2>0$ depend on $N$ and ${\bm G}$.
\end{lmm}
\begin{proof}
Given (5.17), it is sufficient to estimate $\|{\bm B}^{(n)}\|$.
From the definition and Lemma 5.2, we have 
\begin{align*}
&\| {\bm B}^{(n)} \| \leq \sum_{j=1}^N \sum_{l=0}^n
   \bigl( A_1 B_1^{l-1 }\bigr) \bigl( A_1 B_1^{n-l}\bigr) 
 = A_1^2 B_1^{n-1}nN.
\end{align*}
As $n \leq {\rm e}^n \; (n=1,2,\ldots),$ we have
$nB_1^{n-1} \leq {\rm e} ({\rm e}B_1)^{n-1}$, yielding the desired estimate.
\end{proof}
\begin{lmm}
Assume {\rm (3.3)} holds. Then, for $n=1,2,\ldots,$ we have
\begin{align}
&\| {\bm T}^{(n)} \| \leq A_3B_3^{n-1} \; (n=1,2,\ldots),
\end{align}
where the constants $A_3,\; B_3>0$ depend on $N$ and ${\bm G}$.
\end{lmm}
\begin{proof}
Recalling Eq. (5.1) in Subsection 5.1, and following the discussions in Lemma 5.1, we obtain an estimate of the form:
\begin{align*}
&|I_{\tau(n)}^{(i,j)}| \leq A_4 B_4^{n-1} \quad (\tau=1,3).
\end{align*}
Likewise, we have 
\begin{align*}
&\Biggl|
 \sum_{r=1}^k \frac{ \;_\frac12 C_r}{g_{ij}^r} \sum
  \Bigl( \prod_{\tau=1}^r g_{jk}^{p_\tau}\Bigr)
   \Biggr|
    \leq
     A_5 B_5^{k-1}, \quad
%
\Biggl|
 \sum_{r=1}^{n-k} \frac{ \;_\frac12 C_r}{g_{ij}^r} \sum
  \Bigl( \prod_{\tau=1}^r g_{jk}^{p_\tau}\Bigr)
   \Biggr|
 \leq
 A_5 B_5^{n-k-1}, 
\end{align*}
from which we obtain
\begin{align*}
&|I_{2(n)}^{(i,j)}| \leq 
 |t_{ij}| \sum_{k=1}^{n-1} A_5^2B_5^{n-2} 
  = | t_{ij}| (n-1) A_5^2 B_5^{n-2}.
\end{align*}
As described in the proof of Lemma 5.3, this is estimated by $A_5 \widetilde{B}^{n-1}$ with sufficiently large $\widetilde{B}$.
\end{proof}

\begin{lmm}
Assume {\rm (3.3)} holds. Then, for $n=1,2,\ldots,$ we have 
\begin{align*}
&\| \hat{\bm P}_h^{(n)} \| \leq A_6 B_6^{n-1} \; (h=1,2,\ldots ,s),
\end{align*}
where the constants $A_6,\; B_6>0$ depend on $N$, elements of ${\bm G}$, and the isolation distance of ${\bm T}$.
\end{lmm}
\begin{proof}
From (5.20), we have 
\begin{align*}
&\|\hat{\bm P}_h^{(n)}\| \leq \frac{1}{2\pi}
 \sum_{\nu_1+ \ldots + \nu_p = n \atop \nu_1, \ldots, \nu_p \geq 1}
 \prod_{r=1}^p \bigl\| {\bm U}^{(\nu_r)} \bigr\| 
 \int_{ \hat{\Gamma}_h} \bigl\| \hat{\bm R} (\zeta) \bigr\|^{p+1}
 \; {\rm d}\zeta,
\end{align*}
where $\hat{\Gamma}_h$ is introduced in (5.20), and the summation on the right-hand side is taken over all possible values of 
$p \in {\bm N}$ and $(\nu_1, \ldots, \nu_p)$ meeting the condition below the summation symbol. By the normality of ${\bm U}$, we have
\begin{align}
&\bigl\| \hat{\bm R}(\zeta) \bigr\| = \frac{1}{{\rm dist} (\zeta, \sigma({\bm U}))}
\quad \forall \zeta \in P({\bm U}).
\end{align}
Note that in this case, the convergence radius $r_h$ of the series (5.18) and (5.19) is lower-bounded by 
\begin{align*}
&r_h = \Bigl( \frac{2A_2}{ \hat{d}_h} + B_2 \Bigr)^{-1},
\end{align*}
where $A_2$ and $B_2$ are given in Lemma 5.3, and $\hat{d}_h$ is the isolation distance of $\mu_h$. Taking $\hat{\Gamma}_h$ as a circle of radius $\hat{d}_h/2$ surrounding $\mu_h$, we have 
$$
\bigl\| \hat{\bm R}(\zeta) \bigr\| = 2/\hat{d}_h \quad \forall \zeta \in \hat{\Gamma}_h.
$$
Then,
\begin{align*}
&\int_{\hat{\Gamma}_h}
  \bigl\| \hat{\bm R}(\zeta) \bigr\|^{p+1} \leq 
   \pi \hat{d}_h \times \bigl( 2/\hat{d}_h \bigr)^{-(p+1)}
    = 2^{-(p+1)} \pi \hat{d}_h^{-p}.
\end{align*}
Thus, we have 
\begin{align*}
&\bigl\| \hat{\bm P}_h^{(n)} \bigr\| \leq 
  B_2^n 2^{-2}
  \sum_{\nu_1+ \ldots + \nu_p = l \atop \nu_1, \ldots, \nu_p \geq 1} 
    \bigl( A_2/2B_2 \hat{d}_h \bigr)^p
\end{align*}
Note that, from the relationship
$
\mu_h = \lambda_h \pm {\rm i} \sqrt{ 1 - \lambda_h^2},
$
the isolation distance $d_h$ is estimated from below
by that of the corresponding $\lambda_h$.
Now, let us consider the summation on the right-hand side. 
This sum is taken over all possible values of $p$ and $(\nu_1, \ldots, \nu_p)$ meeting the condition below the summation symbol, meaning that $p$ runs from $1$ to $l$.
For fixed $p \; (1 \leq p \leq l)$, the number of sets $(\nu_1, \ldots, \nu_p)$ satisfying
$\nu_1+ \ldots + \nu_p = l$ and $\nu_1, \ldots, \nu_p \geq 1$ is 
$\;_{l-1} C_{l-p}$. Thus, defining $c_0 \equiv  \bigl( A_2/2B_2 \hat{d}_h \bigr)$, we have 
\begin{align*}
&B_2^n \sum_{\nu_1+ \ldots + \nu_p = l \atop \nu_1, \ldots, \nu_p \geq 1} c_0^{-p}
 = (B_2/c_0)^n c_0^n 
   \sum_{p=1}^l \;_{n-1} C_{n-p} c_0^{-p} \\
&\hspace{27mm}
 = (B_2/c_0)^n (1+c_0)^{n-1},
\end{align*}
leading to an estimate of the desired form.
\end{proof}
The following lemma naturally follows by combining (5.26)--(5.27) and Lemma 5.5.
\begin{lmm}
Assume {\rm (3.3)} holds. Then, for $n=1,2,\ldots,$ we have 
\begin{align*}
&\bigl\| {\bm Q}^{(n)} \bigr\| \leq A_7 B_7^{n-1},
\end{align*}
where the constants $A_7, \; B_7>0$ depend on $N$, elements of ${\bm G}$, and the isolation distance of ${\bm T}$.
\end{lmm}
\begin{proof}
As ${\bm U}$ is normal, $\|\hat{\bm P}_h\|=1$ for $h=1,2,\ldots,s$. By virtue of
(5.26)--(5.27), we then have $  \|{\bm Q} \| \leq s$ and
\begin{align*}
%
&\|{\bm Q}^{(n)}\| \leq 
 \sum_{h=1}^s \Biggl[
   \sum_{r_1=1}^r (n-r_1+1)  \|\hat{\bm P}_h^{(r_1)} \| \|\hat{\bm P}_h^{(n-r_1+1)} \| 
%
 +   (n+1) \|\hat{\bm P}_h^{(n+1)} \|
\Biggr] \quad (n=1,2,\ldots).
\end{align*}
From Lemma 5.5, we then have
\begin{align*}
&\hspace{40mm}
 \|\hat{\bm P}_h^{(r+1)} \| \leq A_6 B_6^r, \\[5pt]
&\|\hat{\bm P}_h^{(r_1)} \|\| \hat{\bm P}_h^{(r-r_1+1)} \| \leq A_6 B_6^{r_1-1} \times A_6B_6^{r-r_1}
 = A_6^2B_6^{r-1} \quad  (r_1=1,2,\ldots,r), \\[5pt]
&\hspace{26mm}
 \|\hat{\bm P}_h^{(r)} \|\| \hat{\bm P}_h^{(1)} \| \leq A_6B_6^{r-1} \times A_6 
 = A_6^2B_6^{r-1}, 
\end{align*}
and therefore
\begin{align*}
&\|{\bm Q}^{(n)}\| \leq \sum_{h=1}^s
 \Biggl[
    (n+1) A_6B_6^n  + A_6^2B_6^{n-1} 
%
+ \sum_{r_1=1}^{n-1} (n-r_1+1) A_6B_6^{n-1}
 \Biggr],
\end{align*}
leading to the desired estimate.
\end{proof}
Lemma 5.7 below then follows from Lemma 5.6. 
\begin{lmm}
Assume {\rm (3.3)} holds. Then, for $n=1,2,\ldots,$ we have 
\begin{align*}
&\bigl\| \ket{ \mu^{(n)}} \bigr\| \leq A_8 B_8^{n-1} \quad (n=1,2,\ldots),
\end{align*}
where the constants $A_0,\; B_0>0$ depend on $N$, elements of ${\bm G}$, and the isolation distance of ${\bm T}$.
\end{lmm}
\begin{proof}
From (5.28) and given that $\| \ket{\mu} \|=1$, it is sufficient to show that 
\begin{align*}
&\bigl\| {\bm V}^{(n)} \bigr\| \leq A_9 B_9^{n-1} \quad (n=1,2,\ldots). 
\end{align*}
The desired estimate is found by induction. Assuming $\|{\bm V}^{(j)} \| \leq A_9B_9^{j-1}$ for $j=2,3,4,\ldots,n-1$, and noting that
$\|{\bm Q}\| \leq s$, we have
\begin{align*}
&\|{\bm V}^{(n)} \| \leq 
 \frac{1}{n} \Bigl\{ s+ A_7A_9 \sum_{j=2}^{n-2} B_7^{j-1}B_9^{n-2-j} \Bigr\} \\
&\hspace{10mm}
 \leq s + A_7A_9 \bigl( B_7+B_9\bigr)^{n-1}.
\end{align*}
Setting $\bar{B} = \max\{1, \bigl( B_7+B_9\bigr) \}$, we have 
\begin{align*}
&\|{\bm V}^{(n)} \| \leq s\bar{B} + A_7A_9 \bar{B}^{n-1} 
    \leq \bigl( s+ A_7A_9 \bigr) \bar{B}^{n-1},
\end{align*}
which is the desired estimate.  As $\| \ket{\nu}\|=1$, the estimate of $ \| \ket{ \nu^{(n)}} \|$ is the same as the above.
\end{proof}
\subsection{Estimate of $|N_q^{(n)} (i,m|\psi(0))| $ }
We now estimate $|N_q^{(n)}(i,m|\psi(0)) |$ for $n=1,2,\ldots$. 
\begin{lmm}
Assume {\rm (3.3)} holds. Then, for $n=1,2,\ldots,$ we have 
\begin{align*}
&| N_q^{(n)}(i,m|\psi(0)) | \leq A_{11}B_{11}^{n-1},
\end{align*}
where the constants $A_{11},\; B_{11}>0$ depend on $N$, $m$, and on
elements and the isolation distance of ${\bm G}$.
\end{lmm}
\begin{proof}
Recall the representation (5.29).
From Lemma 5.7, $\| \ket{ \psi(0)} \| = 1$,
and given that $\ket{j,i}$ forms an orthonormal basis in ${\mathcal H}$, we have
\begin{align*}
& \Bigl| \braket{ \mu^{(r_2)} | \psi(0)} \Bigr| \Bigl| \braket{j,i | \mu^{(n-r_1-r_2)}} \Bigr| 
 \leq A_8^2 B_8^{n-r_1-2}.
\end{align*}

To estimate $|\mu^{(n)}|$, we recall that 
$$
\mu^{(n)} = \Bigl( \frac{{\rm d}}{{\rm d}\chi} \Bigr)^n \mu(\chi) \Biggr|_{\chi=0}
 = \lambda^{(n)}
   + 
    \frac{ {\rm i} }{n!}
 \Bigl( \frac{{\rm d}}{{\rm d}\chi} \Bigr)^n 
 \Bigl( 1-(\lambda(\chi))^2 \Bigr)^\frac12
 \Biggr|_{\chi=0}. 
$$
We estimate the second term of the right-most equality in the above expression. By Eq. (5.8), Remark 5.1, and using $r_0$ in Subsection 5.5,
we have $|\lambda^{(p_j)} | \leq \varrho_0 r_0^{p_j} \; (j=1,2,\ldots,r)$.
We also note that $p_1^\prime ! \ldots p_{r^\prime}^\prime ! \geq 1$. We then have 
\begin{align}
&\Bigl| \Bigl( 1-(\lambda(\chi))^2 \Bigr)^\frac12  \Bigr| \leq  r_1^{-n} \sum_{r=1}^n
\varrho_1^r \;_{n-1}C_{r-1} 
%
  \sum_{r^\prime=1}^r
   2^{r^\prime} \;_{r-1}C_{r^\prime-1}
   \bigl|
    \;_\frac12 C_{r^\prime} 
   \bigr|
    \bigl( 1-\lambda^2 \bigr)^{-\frac{(2r^\prime-1)}{2}}.
\end{align}
Letting
$|\lambda |^2 < 1-\varepsilon_1,$
and defining $\varepsilon_1^{-1} = \delta_1 \times (\delta_1 \varepsilon_1)^{-1}$ with some  
 $\delta_1 \in (0,1)$, we can write
$$
|1-\lambda^2|^{-r^\prime} < \delta_1^{r^\prime}  (\delta_1 \varepsilon_1)^{-r^\prime}.
$$
We then have
\begin{align*}
&\sum_{r^\prime=1}^r \bigl(2/(\delta_1\varepsilon_1) \bigr)^{r^\prime}
\;_{r-1}C_{r^\prime-1}
   \bigl|
    \;_\frac12 C_{r^\prime} 
   \bigr|
 \delta_1^{r^\prime} 
%
 \leq 
 \Biggl(
  \sum_{r^\prime=1}^r \bigl(2/(\delta_1\varepsilon_1) \bigr)^{r^\prime}
   \;_{r-1}C_{r^\prime-1}
 \Biggr)
 \Biggl( 
   \sum_{r^\prime=1}^r
   \bigl|
    \;_\frac12 C_{r^\prime} 
   \bigr|  \delta_1^{r^\prime} 
  \Biggr) \\
&\hspace{55mm}
 = \bigl(2/(\delta_1\varepsilon_1) \bigr)
  \Biggl\{
   1+\bigl(2/(\delta_1\varepsilon_1) \bigr)
  \Biggr\}^{r-1}
 \Bigl\{ 
   1 - \Bigl( 1-\delta_1 \Bigr)^\frac12
 \Bigr\} \\
&\hspace{55mm}
 \equiv A_{12}B_{12}^r.
\end{align*}
From these expressions, we can estimate (5.35). We have 
\begin{align*}
&\Bigl| \Bigl( 1-(\lambda(\chi))^2 \Bigr)^\frac12  \Bigr| 
 \leq 
 A_{12}
 r_1^{-n} \sum_{r=1}^n
 (\varrho_1B_{12})^r \;_{n-1}C_{r-1} \\
&\hspace{26mm}
 = A_{12}B_{12}
    r_1^{-n} \varrho_1 
    \Bigl( 1 + \varrho_1 B_{12} \Bigr)^{n-1}.
\end{align*}
Combining this expression with the estimate of $|\lambda^{(n)} |$, we obtain
the estimate $|\mu^{(n)}| \leq A_{13}B_{13}^{n-1}$
with constants $A_{13}, \; B_{13} > 0.$ Actually, recalling (5.31), and
noting that $|\mu|=1$, we obtain
\begin{align*}
&|a_1(n;m)| \leq \sum_{r=1}^{\min\{2m,n\}} \;_{2m}C_r A_{13}^rB_{13}^{n-r}
  \;_{n-1} C_{n-r} \\
&\hspace{15mm}
 \leq 
 B_{13}^n \Bigl( \sum_{r=1}^{2m} \;_{2m}C_r \Bigr)
 \Bigl( \sum_{r=1}^{n} \;_{n-1}C_{n-r} \bigl( A_{13}/B_{13} \bigr)^r \Bigr) \\
&\hspace{15mm}
 =
 B_{13}^n 2^{2m} \bigl( A_{13}/B_{13} \bigr) 
 \Bigl( 1 + \bigl( A_{13}/B_{13} \bigr) \Bigr)^{n-1},
\end{align*}
which yields an estimate of the form 
$$
|a_1(n;m)| \leq A_{14}B_{14}^{n-1}
$$
with constants $A_{14}, \; B_{14} > 0.$
Thus, we have 
\begin{align*}
&|N_q^{(n)}(i,m|\psi(0))| \leq
   \sum_{\mu, j} \sum_{0 \leq r_1+r_2 \leq n}
    A_{14} B_{14}^{r_1-1} A_8^2 B_8^{n-r_1-2} \\
&\hspace{27mm}
 \leq n A_8^2 A_{14} B_8^{n-2} B_{14}^{-1} \sum_{r_1=0}^n \bigl( B_{14}/B_8 \bigr)^{r_1} \\
&\hspace{27mm}
 \leq nA_{11}B_{14}^n,
\end{align*}
where $A_{11} = A_8^2A_{14}B_8^{-2} / (B_{14}-B_8)$. Without losing generality, we assume that $B_8 > B_{14}$ and again assume a sufficiently large constant $B_{11}>0$
(in case $B_8< B_{14}$, we have a similar estimate).
The right-most-hand side in the above expression is then estimated from above by $A_{11}B_{11}^{n}$.
\end{proof}

We now estimate $|I_q^{(n)} (i,m|\psi(0))|$.
By Lemma 5.8, we have \\
$|N_q(i,m|\psi(0)) | \leq A_{11}$ and
 $|N_q^{(l)}(i,m|\psi(0)) | \leq A_{11}B_{11}^{l-1} \; (l=1,2,\ldots)$. Together with (5.32),
this gives
\begin{align*}
&|I_q^{(n)} (i,m | \psi(0)) | \leq \{2B_{11} + (n-1) \}A_{11}^2B_{11}^{n-2} \quad (n \geq 2).
\end{align*}
Note that  $\{ 2B_{11} + (n-1) \}^{\frac{1}{n-1}} \leq \exists c_0$ where $c_0$ is some 
positive constant. Actually, without loss of generality, we can assume $B_{11}>1$ 
and 
\begin{align*}
&\log \{ 2B_{11} + (n-1) \}^{\frac{1}{n-1}}  = 
 \frac{1}{n-1} \log \{ 2B_{11} + (n-1) \}.
\end{align*}
However, $\log \{ 2B_{11} + (n-1) \} \leq 2B_{11}+n-2$, meaning that 
$\log \{ 2B_{11} + (n-1) \}^{\frac{1}{n-1}} $ is bounded.
Accordingly, $\{ 2B_{11} + (n-1) \}^{\frac{1}{n-1}}$ is also bounded, which
 completes the proof of Theorem 4.1.

\section{Conclusion}
This paper discussed the sensitivity analysis of quantum PageRank proposed in
\cite{Paparo_2012}. 
We analytically showed that perturbing the original Google matrix alters the temporal value of quantum PageRank. In addition, we found the lower limit of the convergence radius of the expansion of the perturbed PageRank. 
On the other hand, the temporally averaged value of quantum PageRank and its temporal limitation did not analytically depend on the perturbation in general. 
In future study, we will extend the definition of quantum PageRank and the sensitivity analysis to graph limits (called graphons). 
%


\end{document}